\documentclass{ieeeaccess}
\usepackage{cite}
\usepackage{amsmath,amssymb,amsfonts}
\usepackage{algorithmic}
\usepackage{graphicx}
\usepackage{textcomp}
\usepackage{hyperref}
\usepackage{color}
\usepackage{amssymb}
\usepackage{rotating}
\usepackage[T1]{fontenc}
\usepackage{lscape}
\usepackage{algorithmic}% http://ctan.org/pkg/algorithms
\usepackage{dblfloatfix}
\usepackage{xcolor}
\usepackage{subcaption}
\usepackage[ruled,vlined]{algorithm2e}
\usepackage{float}
\usepackage{tabularx}
\usepackage{ragged2e}
\usepackage{tabulary,booktabs}

\usepackage{multicol}
\usepackage{adjustbox}
\usepackage{csquotes}
\usepackage{longtable}
\usepackage{footnote}

\usepackage{multicol}
\usepackage{float}
\usepackage{amssymb}% http://ctan.org/pkg/amssymb
\usepackage{pifont}% http://ctan.org/pkg/pifont

\DeclareUnicodeCharacter{2212}{-}

\usepackage{etoolbox} % Required for the following command
\makeatletter
\patchcmd{\fnum@table}{\thetable}{\Roman{table}}{}{}
\makeatother

\def\thevol{X}
\def\myyear{XXXX}
\makeatletter
\patchcmd{\@evenfoot}{2016}{\myyear}{}{}
\patchcmd{\@oddfoot}{2016}{\myyear}{}{}
\makeatother

\def\BibTeX{{\rm B\kern-.05em{\sc i\kern-.025em b}\kern-.08em
    T\kern-.1667em\lower.7ex\hbox{E}\kern-.125emX}}
\begin{document}
\history{Date of publication xxxx 00, 0000, date of current version xxxx 00, 0000.}
\doi{XX.XXXX/ACCESS.XXXX.DOI}

\title{Enhancing Plant Disease Detection: A Novel CNN-Based Approach with Tensor Subspace Learning and HOWSVD-MDA}
\author{\uppercase{Abdelmalik Ouamane}\authorrefmark{1,2},
\uppercase{Ammar Chouchane\authorrefmark{2,3}, \uppercase{Yassine Himeur} }\authorrefmark{4},
\IEEEmembership{Senior Member, IEEE}, \uppercase{Abderrazak Debilou} \authorrefmark{1,2}, \uppercase{Abbes Amira} \authorrefmark{5,6}, \IEEEmembership{Senior Member, IEEE}, \uppercase{Shadi Atalla} \authorrefmark{4}, \IEEEmembership{Senior Member, IEEE}, \uppercase{Wathiq Mansoor} \authorrefmark{4}, \IEEEmembership{Senior Member, IEEE}
and \uppercase{Hussain Al Ahmad} \authorrefmark{4}, \IEEEmembership{Life Senior Member, IEEE}
}

\address[1]{Laboratory of LI3C, University of Biskra, Algeria}
\address[2]{Agence Thématique de Recherche en Sciences de la Santé (ATRSS), Algeria}
\address[3]{University Center of Barika. Amdoukal Road, Barika, 05001, Algeria}
\address[4]{College of Engineering and Information Technology, University of Dubai, Dubai, UAE}
\address[5]{Department of Computer Science, University of Sharjah, UAE; and}
\address[6]{Institute of Artificial Intelligence, De Montfort University, Leicester, United Kingdo}

% \markboth
% {Author \headeretal: Vehicle-to-Vehicle Wireless Charging: The Next
% Wave of Innovation in Electrical Vehicles}
% {Author \headeretal: Vehicle-to-Vehicle Wireless Charging: The Next
% Wave of Innovation in Electrical Vehicles}

\corresp{Corresponding author: Y. Himeur (e-mail: yhimeur@ud.ac.ae).}

\begin{abstract}
Machine learning has revolutionized the field of agricultural science, particularly in the early detection and management of plant diseases, which are crucial for maintaining crop health and productivity. Leveraging advanced algorithms and imaging technologies, researchers are now able to identify and classify plant diseases with unprecedented accuracy and speed. Effective management of tomato diseases is crucial for enhancing agricultural productivity. The development and application of tomato disease classification methods are central to this objective. This paper introduces a cutting-edge technique for the detection and classification of tomato leaf diseases, utilizing insights from the latest pre-trained Convolutional Neural Network (CNN) models. We propose a sophisticated approach within the domain of tensor subspace learning, known as Higher-Order Whitened Singular Value Decomposition (HOWSVD), designed to boost the discriminatory power of the system. Our approach to Tensor Subspace Learning is methodically executed in two phases, beginning with HOWSVD and culminating in Multilinear Discriminant Analysis (MDA). The efficacy of this innovative method was rigorously tested through comprehensive experiments on two distinct datasets, namely PlantVillage and the Taiwan dataset. The findings reveal that HOWSVD-MDA outperforms existing methods, underscoring its capability to markedly enhance the precision and dependability of diagnosing tomato leaf diseases. For instance, up to 98.36\% and 89.39\% accuracy scores have been achieved under PlantVillage and the Taiwan datasets, respectively.
\end{abstract}


\titlepgskip=-15pt

\maketitle

\section{Introduction}
\label{sec:introduction}
\begin{figure*}[t!]
\centering
\includegraphics[width=1\textwidth]{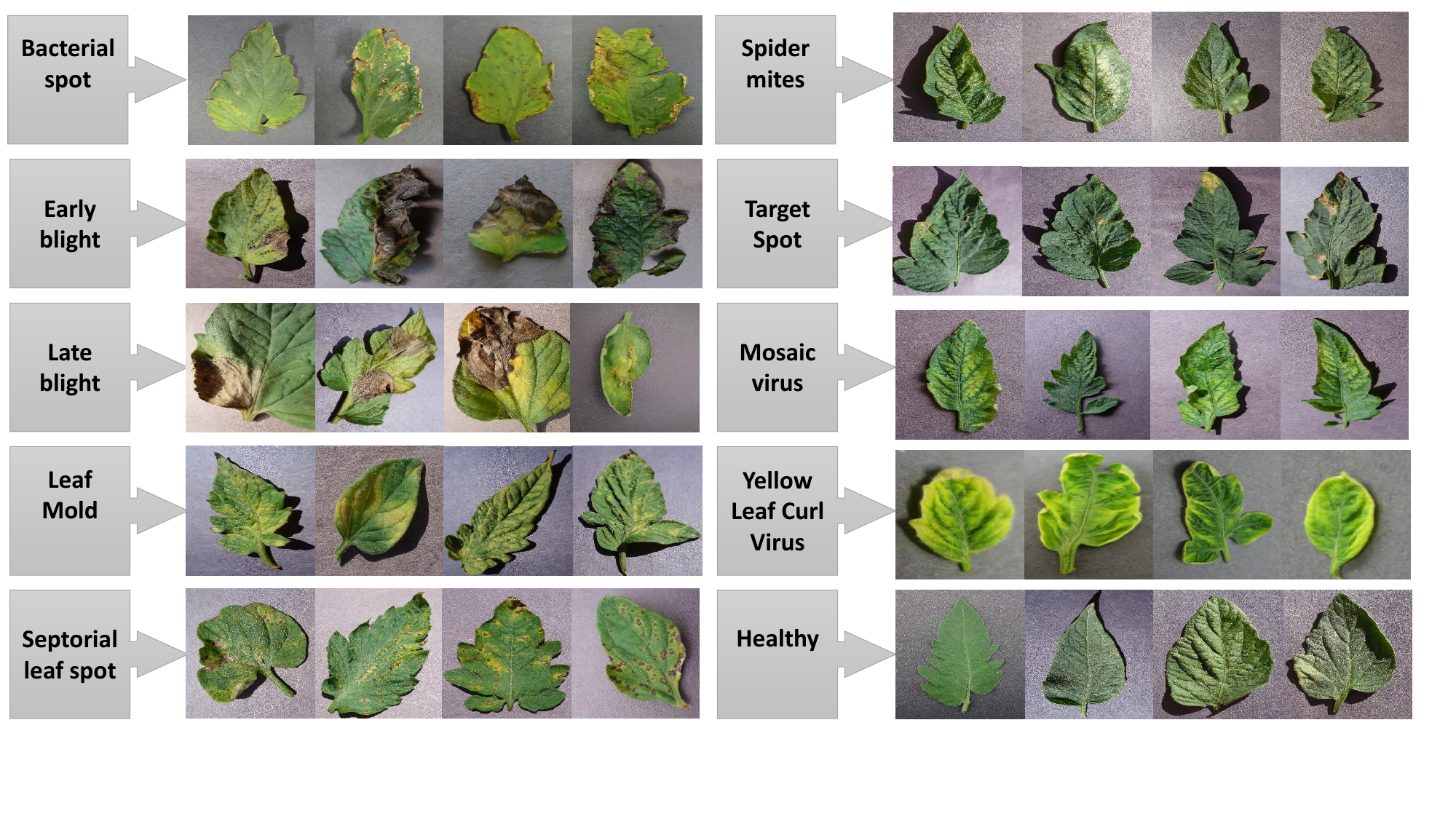}
\caption{Samples of tomato leaves images from PlantVillage dataset}
\label{fig5}
\end{figure*}

Plant crop diseases pose significant risks to global agriculture, threatening food security and causing substantial economic losses \cite{ghofrani2022knowledge}. The rapid identification and classification of these diseases using machine-based approaches remain an active research field \cite{abbas2021tomato}. Notably, Artificial Intelligence (AI) and the Internet of Things (IoT) are increasingly pivotal in agriculture, bolstering crop production and mitigating economic setbacks, particularly concerning plant diseases \cite{atalla2023iot,subeesh2021automation}.
Tomato, a globally cultivated vegetable crop, is notably vulnerable to various diseases throughout its growth stages, including bacterial spots, mosaic virus, late blight, leaf mold, septorial leaf spots, and more \cite{bora2023detection}. These diseases manifest as distinct changes in leaf color and shape, potentially resulting in substantial yield losses if left untreated \cite{nandhini2021improved}. Fig.~\ref{fig5} showcases samples of nine prevalent tomato leaf diseases alongside healthy leaf images sourced from the PlantVillage dataset \cite{hughes2015open}.  

The field of computer vision has witnessed significant advancements in the classification and prediction of plant diseases based on leaf images \cite{himeur2022using}. Deep learning (DL), coupled with image processing, has enabled accurate and automatic systems for tomato leaf disease classification \cite{zampokas2023residual}. Such systems empower farmers to swiftly identify and address plant health concerns, emphasizing high accuracy and reasonable recognition speed to expedite interventions and disease containment \cite{ahmad2023survey}. Multilinear Subspace Learning (MSL), relying on multi-dimensional data representations in the form of high-order tensors, proves instrumental in modeling and analyzing multi-mode or multi-dimensional data, distinguishing it from traditional machine learning (ML) techniques primarily handling single-mode data like vectors \cite{chug2023novel}.

The classification and prediction of plant diseases based on leaf images have become a fertile field for computer vision researchers \cite{chen2023stacking}. Recent advances in computer vision, driven by DL, have opened the door to plant disease diagnosis using leaf images \cite{haridasan2023deep}. This article aims to propose an automatic and accurate system for classifying tomato leaf diseases, allowing farmers to promptly identify and address plant health issues. The system is designed for high accuracy and reasonable recognition speed to expedite interventions and mitigate disease spread in its early stages. Multidimensional data representation through high-order tensors is a powerful tool for Multilinear Subspace Learning (MSL), a critical ML task focusing on analyzing and modeling data with multiple modes or dimensions \cite{makantasis2021rank}. Unlike traditional ML techniques, which typically handle single-mode data like vectors or matrices, multilinear subspace learning deals with data represented as high-order tensors \cite{chouchane2024multilinear}. In this direction, tensors enable the representation and analysis of high-dimensional datasets by breaking them down into components \cite{lu2011survey,ouamane2017efficient}. This approach allows for more efficient identification of data patterns compared to traditional methods like linear or logistic regression. By employing tensor decomposition techniques, we can reduce the dimensionality of large datasets while preserving essential features and relationships between variables. Notably, various computer vision applications have benefited from the strength of this process, including image and video-based biometric recognition \cite{bessaoudi2021multilinear,chouchane2023new}, spectral analysis \cite{deng2023t}, social network analysis \cite{thangamani2021effective}, and biomedical data analysis \cite{chatzichristos2019blind}.

Overall, this study leverages MSL to address the challenges that traditional ML methods often face in capturing complex relationships and interactions among the multiple dimensions inherent in tomato leaf disease images. To our knowledge, this is the first work that combines CNN with MSL, providing potent tensor decomposition and factorization techniques that enhance the discriminative power of pre-trained CNN models. Overall, the main contributions of this paper can be summarized as follows:

\begin{itemize}
\item Conducting a comprehensive literature review to identify research gaps.

\item Intrdocuing an innovative tensor subspace learning method named HOWSVD (Higher-Order Whitened Singular Value Decomposition) for the classification of tomato leaf diseases. HOWSVD enhances the system's discrimination capability by learning discriminative features from high-dimensional CNN embeddings.

\item Implementing a two-step subspace projection process, named as HOWSVD-MDA. This process initially operates the unsupervised HOWSVD approach and is subsequently complemented by MDA (Multilinear Discriminant Analysis), a supervised method. This combination enhances the discriminative capabilities of our system.

\item We conduct a thorough comparative analysis of state-of-the-art pre-trained CNN models, including, Resnet101, Inceptionv3, Resnet50, Nasnetlarge, Alexnet, VGG16, VGG19, Darknet53, Shufflenet, Squeezenet, Xception, Densenet201, Efficientnetb0, Efficientnetb1, Efficientnetb2, Googlenet, Inceptionresnetv2, and Resnet18.This analysis helps identify the most effective model for our specific classification task.

\item To validate the effectiveness of the proposed method, we conduct extensive experiments on two datasets, namely PlantVillage and the Taiwan dataset. This rigorous evaluation demonstrates the robustness and real-world applicability of the proposed HOWSVD-MDA approach.

\end{itemize}

The remaining sections of this paper are organized as follows: section \ref{sec:RW} provides a review of related work on methods for plant disease detection and classification, specifically focusing on deep convolutional neural network-based and ML-based approaches. In section \ref{tensor}, we introduce the proposed HOWSVD-MDA subspace learning method. section \ref{sec:Knowpre} offers an overview of our knowledge pre-trained CNN-based tensor subspace learning. section \ref{sec:Experiments} presents the results of our experiments and provides an analysis of these results. Finally, Section \ref{sec:conclusion} concludes the paper.

\section {Related works}
\label{sec:RW}

\begin{figure*}[ht!]
\centering
\includegraphics[width=1\textwidth]{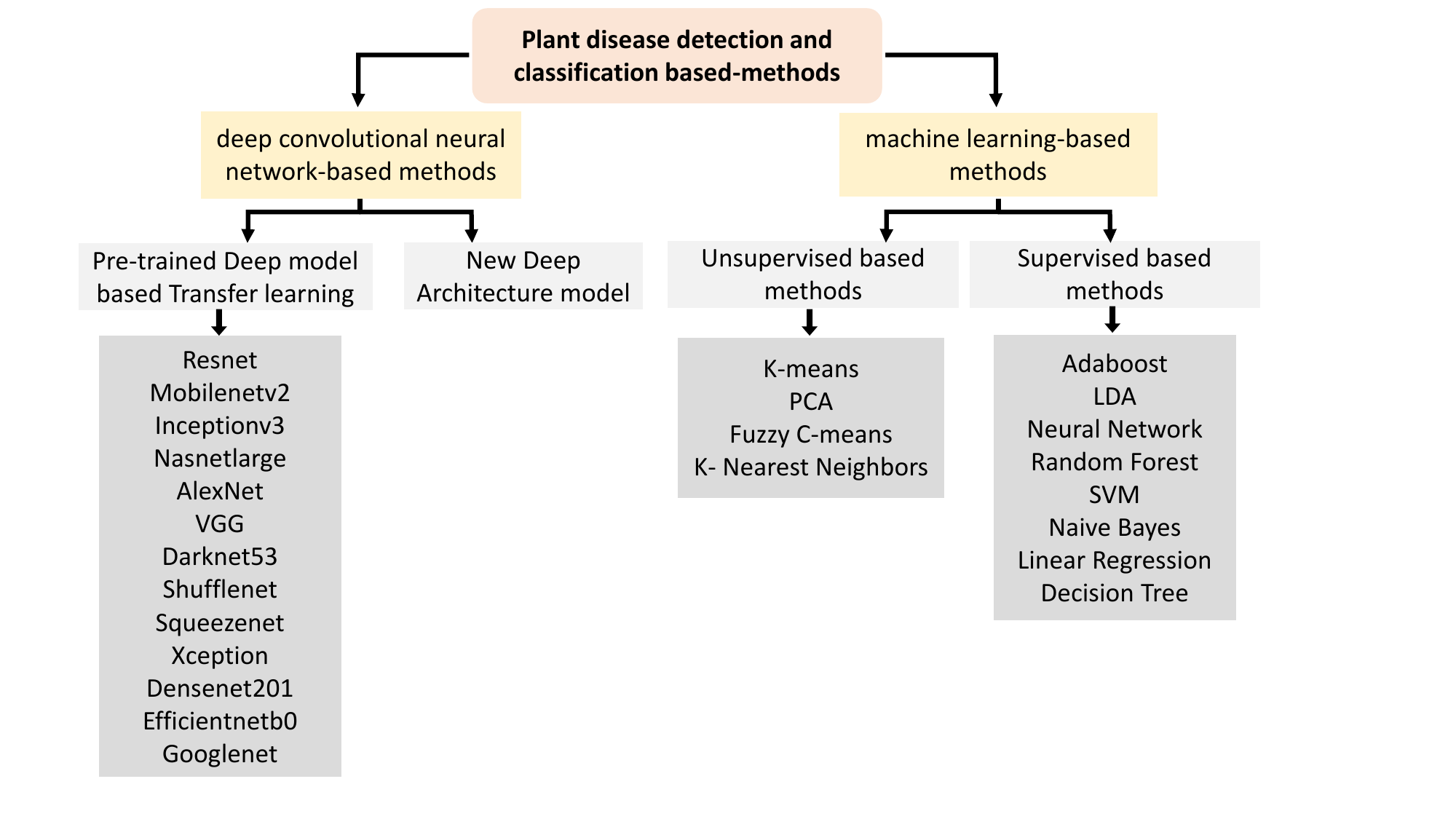}
\caption{Plant disease detection and classification based-methods}
\label{fig1}
\end{figure*}

The development of image-based crop disease identification and classification methods has been a major focus in agricultural computing research \cite{li2023identification}. Typically, these approaches can be classified into two groups: traditional ML-based and deep CNN-based approaches. Both approaches are effective in recognizing diseased crops from images, but they have their advantages and disadvantages. Fig. \ref{fig1} illustrates plant disease classification techniques and highlights well-known algorithms corresponding to each category.
 
Traditional ML-based methods for the classification and detection of plant diseases often rely on visual characteristics such as color, texture, and shape, which are highly sensitive to changes when diseases occur \cite{chen2020identification,tao2014fruits}. These ML-based methods typically consist of two principal components: feature extraction and classification. Feature extraction is employed to identify patterns in input leaf images that are indicative of a particular disease. Features such as color, texture, or shape are extracted before classifying the images using supervised or unsupervised classification algorithms, as illustrated in Fig.~\ref{fig1}.

The feature extraction process in plant disease detection and classification typically involves handcrafted features such as Gabor wavelets transform (GWT) and Gray Level Cooccurrence Matrix (GLCM) \cite{prasad2016multi}, scale-invariant feature transform (SIFT) \cite{patil2017analysis}, Local Binary Pattern \cite{lv2023research}, histogram of oriented gradients (HOG) \cite{tsolakidis2014plant}, and Speeded-Up Robust Features (SURF) \cite{basavaiah2020tomato}. While this approach is relatively simple and straightforward to implement, it often requires a large amount of labeled training data, which may not always be available for certain diseases or regions due to a lack of expertise or resources required for manual labeling tasks. In some recent studies, alternative techniques such as Fisher vector (FV) \cite{kurmi2021leaf} and K-means clustering \cite{kumari2019leaf} have been employed to improve the accuracy of crop disease detection and classification, especially for tomato leaf diseases.

Recently, there has been widespread adoption of deep CNNs models for the classification of plant diseases, including tomato leaf diseases, achieving very high-performance levels \cite{abbas2021tomato,nandhini2021improved,chen2020identification,trivedi2021early,thangaraj2021automated,zhang2022mmdgan}. A typical deep CNN comprises multiple layers, the convolutional layers are responsible for extracting distinctive features from input images by applying filters and detecting patterns. On the other hand, pooling layers help in reducing the size of feature maps obtained from the first step, thereby decreasing computational complexity and aiding in preventing overfitting by summarizing the extracted features. After that, The fully connected layers combine these discriminant features to predict the disease type. CNNs have revolutionized computer vision by autonomously learning complex representations directly from raw data, eliminating the need for manual feature engineering steps. This makes them ideal for capturing subtle patterns in images that may not be easily detected with traditional techniques, resulting in more accurate classifications compared to classical methods such as Support Vector Machines or Random Forests. Furthermore, CNNs demand significantly fewer computational resources than older models, making them particularly attractive for real-time applications like crop disease detection systems deployed at scale across large farms and plantations.

In the following paragraph, we review some of the most significant recent research that relies on deep CNN models for classifying plant diseases, with a focus on tomato leaf diseases. Chen et al. \cite{chen2020identification} propose a novel tomato leaf disease recognition system that employs Binary Wavelet Transform followed by a Retinex filter to eliminate noise effects and extract texture information from leaf images. They utilize the Both-channel Residual Attention Network model (B-ARNet) as the identifying signature for leaf images in their database. In another study, Mingxuan et al. \cite{li2023identification} tackle the challenge of distinguishing between large and small intraclass differences and interclass variations to achieve high accuracy in identifying tomato leaf diseases. They introduce a lightweight neural network model called LMBRNet, which extracts distinctive features from tomato leaf images across various dimensions and receptive fields. To address network degradation during training, they incorporate multiple residual connections across all layers. Additionally, Too et al. \cite{too2019comparative} conduct an insightful comparative study of fine-tuning DL architectures, evaluating VGG16, Inception, ResNet, and DenseNets models on the PlantVillage dataset. Their findings confirm that DenseNets outperform other models in terms of test accuracy. The training of the neural model is implemented using Keras with Theano as the backend.

In a related study, Hammad Saleem et al. \cite{saleem2020image} evaluated various CNN meta-architectures, including SSD, Faster RCNN, and RFCN, on the PlantVillage dataset. They utilized Adam and RMSProp optimizers for the weight parameters, resulting in a notable performance improvement for these models. Brahim et al. \cite{brahimi2017deep} proposed the use of efficient pre-trained DL models like GoogleNet and AlexNet for disease classification in tomato leaves. Their approach involved localizing the infected regions in leaf images using occlusion techniques, aiding users in visualizing and understanding the diseases. The authors compared their deep model with shallow models like SVM and Random Forest, confirming the superior effectiveness of CNN deep models in achieving high classification accuracy. Mariam Moussafir et al. \cite{moussafir2022design} introduced a novel concept using Genetic Algorithms (GAs) for disease detection in tomato leaves. They employed fine-tuning strategies to enhance pre-trained deep models, and GA was used to optimize CNN hyperparameters. Many other research works have adopted various DL approaches for disease detection and classification based on leaf images, including references \cite{abbas2021tomato,bora2023detection,paymode2022transfer,anandhakrishnan2022deep,zhao2022ric,astani2022diverse,gajjar2021real,chug2022novel,nandhini2022automatic,tembhurne2023plant}.

\begin{table*}[t!]
\centering
\caption{Comparison of Studies on Image-Based Crop Disease Identification and Classification}
\label{tab:studyComparison}
\begin{tabular}{@{}p{0.6cm}p{3cm}p{2cm}p{3cm}p{2.5cm}p{4.4cm}@{}}
\hline
\textbf{Ref.} & \textbf{ML Model Used} & \textbf{Dataset} & \textbf{Contribution} & \textbf{Best Performance} & \textbf{Limitation} \\ \hline

\cite{chen2020identification} & B-ARNet & Not specified & Novel system using Binary Wavelet Transform and Retinex filter & High accuracy in noise-affected images & Dataset specifics and scalability not mentioned \\

\cite{li2023identification} & LMBRNet & Not specified & Lightweight model for distinguishing intraclass differences & High accuracy with complex intraclass variations & Detailed performance metrics not provided \\

\cite{too2019comparative} & DenseNets (compared VGG16, Inception, ResNet, DenseNets) & PlantVillage & Comparative study of DL architectures & DenseNets outperformed other models & Limited to tested architectures \\

\cite{saleem2020image} & SSD, Faster RCNN, RFCN & PlantVillage & Evaluated CNN meta-architectures with Adam and RMSProp optimizers & Notable performance improvement & Performance comparison metrics not detailed \\

\cite{brahimi2017deep} & GoogleNet, AlexNet & Not specified & Used pre-trained models for disease classification, employing occlusion & Superior to SVM and Random Forest & Limited discussion on real-world application \\

\cite{moussafir2022design} & Pre-trained models optimized with Genetic Algorithms & PlantVillage & GA for optimizing CNN hyperparameters & accuracy = 98.1\% & No insights about generalization to other crops, real-world deployment challenges, computational requirements, and the risk of overfitting.  \\

\cite{abbas2021tomato} & DenseNet121 & PlantVillage + synthetic images dataset & Adoption of CNN models for plant disease classification & accuracy = 99.51\% & General limitations of CNNs (e.g., data dependency) \\
\hline
\end{tabular}
\end{table*}

%%%%%%%%%%%%%%%%%%%%%%%%%%%%%%%%%%%%%%%%%%%%%%%%%%%%%%%%%%%%%%%%%%%%%%%%%%%%%%%%%%%%%%%%
\section{HOWSVD-MDA Subspace Learning}\label{tensor}

In this section, we introduce the enhanced subspace tensor approach, a pivotal component of our tomato leaf disease classification system. We provide a detailed overview of the materials and formulations that underpin this innovative approach.

\subsection{Materials and Formulations}

The introduction of notations and basic tensor fundamentals is essential to emphasize the mathematical aspect of this paper, which focuses on tensor-based multidimensional image applications. To enhance the readability of the equations presented throughout the paper, we represent each data type using a unique symbol format (scalars, matrices, tensors). Table \ref{tab:T1} lists the operations associated with tensors, a brief summary of the fundamental tensor operations and properties relevant to this paper, and the primary tensor representations.

\subsubsection{Tensor definition}
A tensor is a mathematical entity used for multi-dimensional data representation \cite{ouamane2017efficient}. It is characterized by its rank or order, which determines the number of indices required to access its components. Tensors can be scalars ($0^{th}$-order tensors), vectors ($1^{st}$-order tensors), matrices ($2^{nd}$-order tensors), or higher-order tensors (higher-dimensional arrays) with N entries. Let $\boldsymbol{\mathrm{X}}$ be a tensor of order N, represented as $\mathrm{\boldsymbol{\mathrm{X}}\in {\mathbb{R}}^{I_1 \times I_2 \times \cdots \times I_N}}$. It is described by N indices, where $\mathrm{I_k}$ for $\mathrm{1 \le k \le N}$ indicates the dimension of the k-th mode of the tensor. Each element of the tensor $\boldsymbol{\mathrm{X}}$ is denoted by $\mathrm{x_{\mathrm{i_1}, \mathrm{i_2}, \ldots, \mathrm{i_N}}}$.

\subsubsection{k-Mode Unfolding (Matricization)  }
\begin{figure*}[t!]
\centering
\includegraphics[width=1\textwidth]{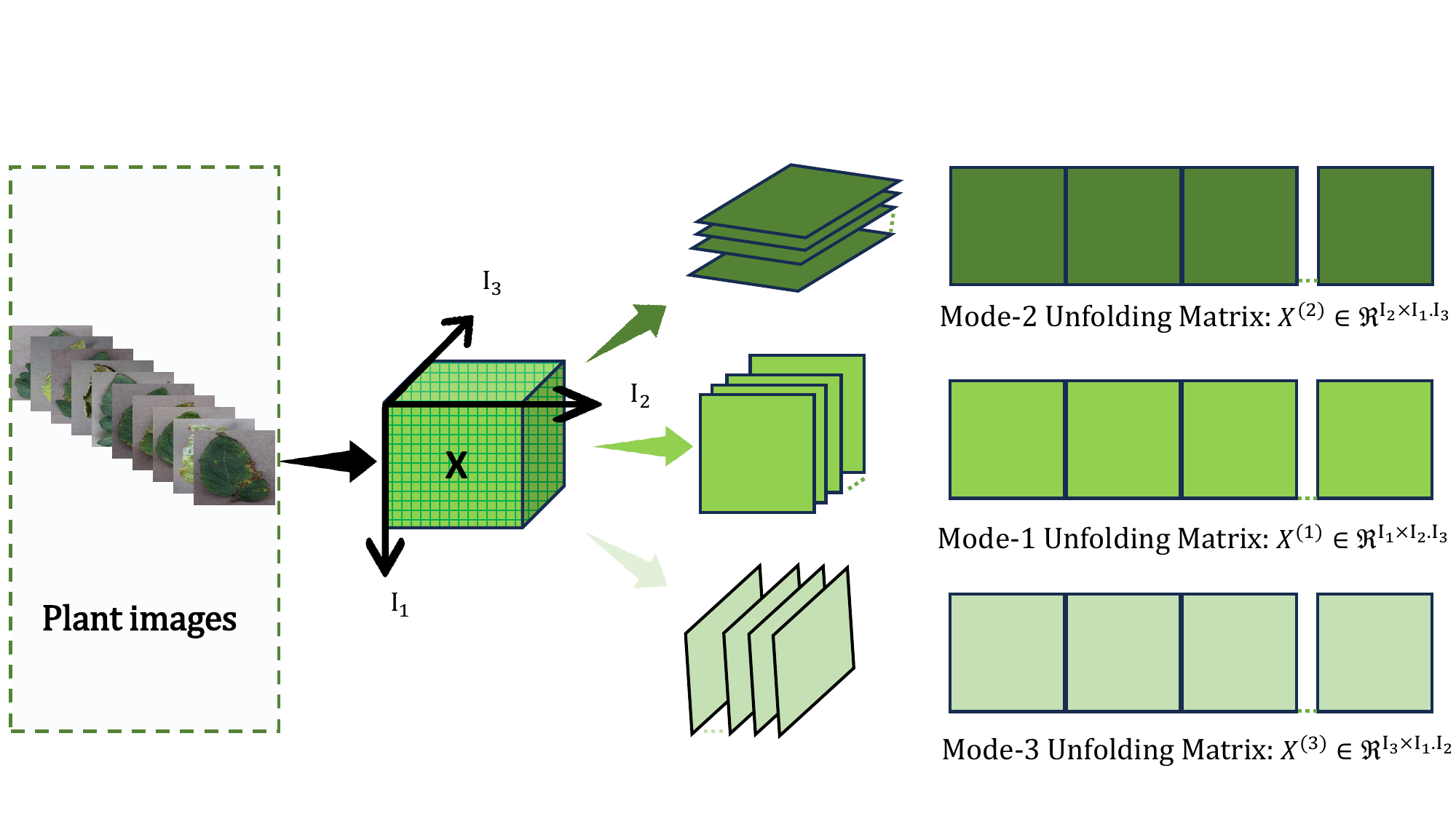}
\caption{Visualization of the 3-Mode Unfolding of a 3$^{th}$-order tensor}
\label{fig2}
\end{figure*}

To efficiently manipulate tensors for multidimensional data representation, it is essential to introduce fundamental tools from multilinear algebra. One critical operation in this context is k-Mode unfolding (see Fig.~\ref{fig2}), also known as matricization. This data transformation operation is frequently used in multidimensional data analysis and tensor factorization to convert a high-order tensor into a matrix, simplifying the data analysis and manipulation process. Thus, the unfolding matrix $X^{\mathrm{(k)}}\mathrm{\in }\mathrm{\ }{\mathbb{R}}^{{\mathrm{I}}_{\mathrm{k}}\mathrm{\times }\prod{\mathrm{i}\mathrm{\neq }\mathrm{k}}_{{\mathrm{I}}_{\mathrm{i}}}}$ in the k-mode of the tensor $\boldsymbol{\mathrm{X}}$ is represented by: $\boldsymbol{\mathrm{X}}{\mathrm{\Rightarrow }}_{\mathrm{k}}X^{\mathrm{(k)}}$, in which: ${{X}}^{\mathrm{(k)}}_{{\mathrm{i}}_{\mathrm{k}}\mathrm{j}}\mathrm{=}{\boldsymbol{\mathrm{X}}}_{{\mathrm{i}}_{\mathrm{1}}\mathrm{\cdots }{\mathrm{i}}_{\mathrm{m}}}, \mathrm{\ \ j=1+}\sum^{\mathrm{m}}_{\mathrm{l=1,l}\mathrm{\neq }\mathrm{k}}{\mathrm{(}{\mathrm{i}}_{\mathrm{l}}\mathrm{-}\mathrm{1)}\prod^{\mathrm{m}}_{\mathrm{o=l+1,o}\mathrm{\neq }\mathrm{k}}{{\mathrm{I}}_{\mathrm{o}}}}$. 

The k-mode unfolding of the tensor allows for a more structured and interpretable representation of high-order tensor data, enabling the application of various analytical techniques, including dimensionality reduction, visualization, clustering, classification, and tensor decomposition \cite{ouamane2017efficient,chouchane2023new,surana2022hypergraph}. It serves as a foundational step in complex multidimensional data analysis and plays a pivotal role in making tensor-based approaches accessible and effective across a wide range of applications. Fig. \ref{fig2} presents a visualization of the k-Mode Unfolding of third-order tensor data.

\subsubsection{k-Mode Product of a tensor}
The k-Mode Product of a tensor refers to a mathematical operation that combines a tensor with a matrix along a specific mode or dimension. This product can be interpreted as a scalar product in the k-mode. The k-mode product of a tensor  $\mathrm{\boldsymbol{\mathrm{X}}\in {\mathbb{R}}^{I_1 \times I_2 \times \cdots \times I_N}}$ by a matrix  $U\boldsymbol{\mathrm{\in }}\boldsymbol{\mathrm{\ }}{\mathbb{R}}^{{\mathrm{I^{'}}}_{\mathrm{K}}\mathrm{\times }{\mathrm{I}}_{\mathrm{k}}}$ $( \mathrm{k=1, 2,\dots, N})$ is denoted by $\boldsymbol{\mathrm{X}}{\mathrm{\times }}_{\mathrm{k}}U$ and gives a tensor $\mathrm{\boldsymbol{\mathrm{Y}}}$ such that: ${\boldsymbol{\mathrm{Y}}}_{{\mathrm{i}}_{\mathrm{1}}\mathrm{,}\mathrm{\cdots }{\mathrm{,i}}_{\mathrm{k-1}}\mathrm{,i,}{\mathrm{i}}_{\mathrm{k+1}}\mathrm{,}\mathrm{\cdots }{\mathrm{,i}}_{\mathrm{N}}}\mathrm{=}\sum^{{\mathrm{I}}_{\mathrm{k}}}_{\mathrm{j}=1}{{\boldsymbol{\mathrm{X}}}_{{\mathrm{i}}_{\mathrm{1}}\mathrm{,}\mathrm{\cdots }{\mathrm{,i}}_{\mathrm{k-1}}\mathrm{,j,}{\mathrm{i}}_{\mathrm{k+1}}\mathrm{, \cdots }{\mathrm{,i}}_{\mathrm{N}}}U_{\mathrm{i,j}}}  $.

\begin{table}[!ht]
    \centering
		   \caption{Main representations in this paper  }
    \begin{tabular}{c|c}
    \hline
        \textbf{Symbol} & Description \\ \hline
				x & Scalar (zero order tensor)  \\ 
        \textit{x} & Vector (one order tensor)  \\ 
        \textit{X} & Matrix (two order tensor)  \\ 
        $\mathrm{\boldsymbol{\mathrm{X}}\in {\mathbb{R}}^{I_1 \times I_2 \times \cdots \times I_N}}$ & High order tensor  \\ 
        ${\mathbb{R}}$ & Real numbers set  \\ 
         
        N & Number of tensor modes  \\ 
        $\mathrm{I_1, I_2, \cdots , I_N}$ & Number of components for each dimension  \\ 
        $\boldsymbol{\mathrm{X}}{\mathrm{\Rightarrow }}_{\mathrm{k}}X^{\mathrm{(k)}}$ & k-mode unfolding of a tensor \textbf{X}  \\ 
        $\boldsymbol{\mathrm{X}}{\mathrm{\times }}_{\mathrm{k}}U$ & k-mode product of a tensor  \\ 
        M & Number of training data  \\ \hline
    \end{tabular}
    \label{tab:T1}
\end{table}

\subsection{Higher-Order Whitened Singular Value Decomposition (HOWSVD)}

The HOSVD \cite{Zarzoso1999}, a multilinear extension of the matrix SVD (Singular Value Decomposition) to high-order tensors, plays a pivotal role across a range of application domains. These include pattern recognition \cite{SERRAOUI2022222,YIN2023109450}, image processing \cite{8756627,2023JEI....32b3028W}, machine learning (ML) and dimensionality reduction \cite{10.1145/1454008.1454017}, signal processing \cite{YUAN2023109923,DELATHAUWER200431}, and biometrics \cite{10.1007/978-3-319-98352-3_23,GUEHAIRIA20221309}. Our research introduces a novel variant, the Higher-Order Whitened Singular Value Decomposition (HOWSVD), which aims to enhance the representation of data tensors.

Consider $\mathrm{\boldsymbol{\mathrm{X}}\in {\mathbb{R}}^{I_1 \times I_2 \times \cdots \times I_N}}$, representing N$^{th}$-order tensor data samples (knowledge feature vectors of the input tomato image). The training sample set is formulated as an (N + 1)$^{th}$-order tensor, $\widetilde{\boldsymbol{\mathrm{X}}} \boldsymbol{\mathrm{\in }}\boldsymbol{\mathrm{\ }}{\mathbb{R}}^{{\mathrm{I}}{\mathrm{1}}\mathrm{\times }{\mathrm{I}}{\mathrm{2}}\mathrm{\times }\mathrm{\cdots }{\mathrm{I}}_{\mathrm{N}}\times \mathrm{M}}$, where M denotes the number of training tomato images. The implementation of the HOSVD algorithm proceeds as outlined:

\begin{itemize} 	
\item For each k mode, compute the covariance matrix: $C^{\mathrm{(k)}}=\sum^{\mathrm{M}}_{\mathrm{i}} X^{\mathrm{(k)}}_{\mathrm{i}} (X^{\mathrm{(k)}}_{\mathrm{i}})^{\mathrm{T}}$,  where $X^{\mathrm{(k)}}_{\mathrm{i}}$ are k-mode unfolding of $\mathrm{\textbf{X}_\mathrm{i}}$. 

\item Compute the SVD for each k covariance matrix $C^{\mathrm{(k)}}: C^{\mathrm{(k)}}=U^{\mathrm{(k)}} \Sigma^{\mathrm{(k)}} {(V^{\mathrm{(k)}})}^\mathrm{T}$,  where $U^{\mathrm{(k)}}$ contains the left singular vectors, $\Sigma^{\mathrm{(k)}}$ denotes the singular value matrices and $V^{\mathrm{(k)}}$ contains the right singular vectors.

\item Let: $U^{\mathrm{(k)}}_{\mathrm{HOSVD}}= [u^{\mathrm{(k)}}_1, u^{\mathrm{(k)}}_2,..., u^{\mathrm{(k)}}_{\mathrm{N}}]$, where $U^{\mathrm{(k)}}_{\mathrm{HOSVD}}$ 	contains the selected left singular vectors corresponding to largest ${\mathrm{N}}_{\mathrm{k}}$ singular value. 

\item The projection of ${\mathrm{\textbf{X}}}_{\mathrm{i}}$ knowledge tensors is realized by the following  k-mode product : 

${\mathrm{\textbf{Y}}}_{\mathrm{i}}={\mathrm{\textbf{X}}}_{\mathrm{i}} \times_1 {(U^{\mathrm{(1)}}_{\mathrm{HOSVD}})}^{\mathrm{T}} \times_2 {(U^{\mathrm{(2)}}_{\mathrm{HOSVD}})}^{\mathrm{T}}...\times_{\mathrm{n}} {(U^{\mathrm{(N)}}_{\mathrm{HOSVD}})}^{\mathrm{T}}$

\end{itemize}

In this paper, we introduce a novel extension of the Higher Order Singular Value Decomposition (HOSVD) method, which we term the Whitened HOSVD (HOWSVD). Our objective is to enhance the tensor-based knowledge representation of tomato image data. 

The key innovation lies in the whitening process, which linearly transforms the data, leading to the covariance matrix becoming the identity matrix. This transformation results in reduced data correlation and uniform variance across all dimensions post-whitening. The whitening effect is consistent across various directions. To achieve this, each eigenvector is normalized by the square root of its corresponding eigenvalue. Extensive experimentation has demonstrated that this preparatory step significantly improves feature discrimination, especially in datasets with high levels of noise.
In the proposed HOWSVD method for tomato image tensor knowledge, the whitening process is applied to each mode of the tensor :
\begin{equation} \label{GrindEQ1} 
 U^{\mathrm{(k)}}_{\mathrm{HOWSVD}}={(\Lambda^{\mathrm{(k)}})}^{-1/2} U^{\mathrm{(k)}}_{\mathrm{HOSVD}}, \Lambda^{\mathrm{(k)}}=[{\mathrm{\lambda}}^{\mathrm{(k)}}_1 {\mathrm{\lambda}}^{\mathrm{(k)}}_2 ...{\mathrm{\lambda}}^{\mathrm{(k)}}_{\mathrm{N}}] 
\end{equation} 

\subsection{HOWSVD-MDA Algorithm}

In this study, we have enhanced the discriminative power of our HOWSVD (Higher-Order Web Services Vector Decomposition) by incorporating an additional level of subspace learning through Multilinear Discriminant Analysis (MDA) \cite{yan2006multilinear}. Building upon the initial multilinear subspace learning applied in our HOWSVD, our approach aims to bolster the discriminative capabilities of tensor-based knowledge representations for tomato images. Specifically, our goal is to increase the inter-class separation and reduce the intra-class variation for each k-mode of the tensor knowledge representation of tomato images, achieved by projecting them in two level, HOWSVD followed by MDA. To accomplish this, we represent the training samples of tomato images as M tensors, $\mathrm{\boldsymbol{\mathrm{X}}\in {\mathbb{R}}^{I_1 \times I_2 \times \cdots \times I_N}}$ belonging to L diverse classes and each class j contains ${\mathrm{n}}_{\mathrm{j}}$ samples. HOWSVD project ${\mathrm{N}}^{\mathrm{th}}$-order tensors $\mathrm{\boldsymbol{\mathrm{X}}\in {\mathbb{R}}^{I_1 \times I_2 \times \cdots \times I_N}}$ to :

\begin{equation} \label{GrindEQ2}
{\mathrm{\textbf{Y}}}_{\mathrm{i}}= {\mathrm{\textbf{X}}}_{\mathrm{i}} \times_1 {(U^{\mathrm{(1)}}_{\mathrm{HOWSVD}})}^{\mathrm{T}} \times_2 {(U^{\mathrm{(2)}}_{\mathrm{HOWSVD}})}^{\mathrm{T}}...\times_{\mathrm{N}} {(U^{\mathrm{(N)}}_{\mathrm{HOWSVD}})}^{\mathrm{T}}
\end{equation} 

Where: $\boldsymbol{\mathrm{Y_i}} \in {\mathbb{R}}^{{\mathrm{I'}}_1 \times {\mathrm{I'}}_2 \times ... \times {\mathrm{I'}}_{\mathrm{N}}}, U^{\mathrm{(k)}}_{\mathrm{HOWSVD}} \in {\mathbb{R}}^{{\mathrm{I}}_{\mathrm{k}} \times {\mathrm{I'}}_{\mathrm{k}}}, {\mathrm{I'}}_1 \prec \prec {\mathrm{I}}_1 , {\mathrm{I'}}_2 \prec \prec {\mathrm{I}}_2 , ..., {\mathrm{I'}}_{\mathrm{N}} \prec \prec {\mathrm{I}}_{\mathrm{N}}   $, (${\mathrm{I'}}_1, {\mathrm{I'}}_2 , ..., {\mathrm{I'}}_{\mathrm{N}}$ the new lower dimensions after HOWSVD method).

MDA \cite{yan2006multilinear} aims to find N projection matrices ${\mathrm{\widetilde{W}}}^{\mathrm{(k)}}$ that are interconnected by maximizing the inter-class scatter while minimizing the intra-class scatter matrices in each k-mode of the $\boldsymbol{\mathrm{Y_i}} \in {\mathbb{R}}^{{\mathrm{I'}}_1 \times {\mathrm{I'}}_2 \times ... \times {\mathrm{I'}}_{\mathrm{N}}}$ tensor:

\textsc{\[{\left.{\mathrm{\widetilde{W}}}^{\mathrm{(k)}}\right|}^{\mathrm{N}}_{\mathrm{k=1}}=\mathop{{\mathrm{arg} \mathrm{max}\ }}_{{\left.W^{\mathrm{(k)}}\right|}^{\mathrm{N}}_{\mathrm{k=1}}}\] 
\begin{equation} \label{GrindEQ3} 
{ \frac{\sum^{\mathrm{L}}_{\mathrm{j=1}}{{\mathrm{n}}_{\mathrm{j}}{\left\|{\overline{\boldsymbol{\mathrm{Y}}}}_{\mathrm{j}}{\times }_1W^{\mathrm{(1)}}\cdots {\times }_{\mathrm{N}}W^{\mathrm{(N)}}-\overline{\boldsymbol{\mathrm{Y}}}{\times }_1W^{\mathrm{(1)}}\cdots {\times }_{\mathrm{N}}W^{\mathrm{(N)}}\right\|}^2}}{\sum^{\mathrm{M}}_{\mathrm{i=1}}{{\left\|{\boldsymbol{\mathrm{Y}}}_{\mathrm{i}}{\times }_1W^{\mathrm{(1)}}\cdots {\times }_{\mathrm{N}}W^{\mathrm{(N)}}-{\overline{\boldsymbol{\mathrm{Y}}}}_{{\mathrm{n}}_{\mathrm{i}}}{\times }_{\mathrm{1}}W^{\mathrm{(1)}}\cdots {\times }_{\mathrm{N}}W^{\mathrm{(N)}}\right\|}^2}} } 
\end{equation}}

Where: ${\overline{\boldsymbol{\mathrm{Y}}}}_{\mathrm{j}}$ is the average tensor of each class $\mathrm{j}$, $\overline{\boldsymbol{\mathrm{Y}}}$ is the average tensor of totally training data.

Eq. \eqref{GrindEQ3} is a problem that requires higher order nonlinear optimization and a higher order nonlinear constraint. Finding a direct closed solution is not noticeable. To estimate the interrelated discriminative subspaces, an iterative optimization approach is used \cite{yan2006multilinear}. We have the following objective function based on the optimization problem of each k-mode: 
\begin{equation} \label{GrindEQ4} 
{\mathrm{\widetilde{W}}}^{\mathrm{(k)}}={\mathop{{\mathrm{arg} \mathrm{max}\ }}_{W^{\mathrm{(k)}}} \frac{\mathrm{Tr}\left({(W^{\mathrm{(k)}})}^{\mathrm{T}}S_{\mathrm{b}}W^{\mathrm{(k)}}\right)}{\mathrm{Tr}\left(({W^{\mathrm{(k)}})}^{\mathrm{T}}S_{\mathrm{w}}W^{\mathrm{(k)}}\right)}\ } 
\end{equation} 

\begin{equation} \label{GrindEQ6} 
S^{\mathrm{k}}_{\mathrm{b}}=\sum^{\mathrm{L}}_{\mathrm{j}=1}{{\mathrm{n}}_{\mathrm{j}}\left({\overline{Y}}^{\mathrm{k}}_{\mathrm{j}}-{\overline{Y}}^{\mathrm{k}}\right){\left({\overline{Y}}^{\mathrm{k}}_{\mathrm{j}}-{\overline{Y}}^{\mathrm{k}}\right)}^{\mathrm{T}}}
\end{equation}

\begin{equation} \label{GrindEQ7} 
S^{\mathrm{k}}_{\mathrm{w}}=\sum^{\mathrm{L}}_{\mathrm{j}=1}{\sum^{{\mathrm{n}}_{\mathrm{j}}}_{\mathrm{i=1}}{{\left(Y^{\mathrm{k}}_{\mathrm{j,i}}-{\overline{Y}}^{\mathrm{k}}_{\mathrm{j}}\right)\left(Y^{\mathrm{k}}_{\mathrm{j,i}}-{\overline{Y}}^{\mathrm{k}}_{\mathrm{j}}\right)}^{\mathrm{T}}}}
\end{equation}

$S^{\mathrm{k}}_{\mathrm{b}}$ and $S^{\mathrm{k}}_{\mathrm{w}}$ are the between and within class scatter matrices, respectively. ${\overline{Y}}^{\mathrm{k}}_{\mathrm{j}}$  is the average matrix on class j, ${\overline{Y}}^{\mathrm{k}}$ is the average matrix of the whole training information and $Y^{\mathrm{k}}_{\mathrm{j,i}}$ is the k-mode unfolded matrix of tensor ${\boldsymbol{\mathrm{Y}}}_\mathrm{i}$ .
Each iteration of the solution of Eq.~\eqref{GrindEQ3} requires the identification of $W^{(1)}, W^{(2)}, ...,W^{\mathrm{N}}$  leading to:

\begin{equation} \label{GrindEQ8} 
  {\boldsymbol{\mathrm{Z}}}_{\mathrm{i}}={\boldsymbol{\mathrm{Y}}}_{\mathrm{i}}{\times }_{\mathrm{i}}W^{\mathrm{(1)}}\cdots {\times }_{\mathrm{k-1}}W^{\mathrm{(k-1)}}{\times }_{\mathrm{k+1}}W^{\mathrm{(k+1)}}\cdots {\times }_{\mathrm{N}}W^{\mathrm{(N)}} 
	\end{equation} 
	
	We find:

\begin{equation} \label{GrindEQ9} 
{\mathrm{\widetilde{W}}}^{\mathrm{(k)}}={\mathop{{\mathrm{arg} \mathrm{max}\ }}_{W_{\mathrm{(k)}}} \frac{\sum^{\mathrm{L}}_{\mathrm{j=1}}{{\mathrm{n}}_{\mathrm{j}}{\left\|{\overline{\boldsymbol{\mathrm{Z}}}}_{\mathrm{j}}{\times }_{\mathrm{k}}W^{\mathrm{(k)}}-\overline{\boldsymbol{\mathrm{Z}}}{\times }_{\mathrm{k}}W^{\mathrm{(k)}}\right\|}^2}}{\sum^{\mathrm{M}}_{\mathrm{i=1}}{{\left\|{\boldsymbol{\mathrm{Z}}}_{\mathrm{i}}{\times }_{\mathrm{k}}W^{\mathrm{(k)}}-{\overline{\boldsymbol{\mathrm{Z}}}}_{{\mathrm{n}}_{\mathrm{i}}}{\times }_{\mathrm{k}}W^{\mathrm{(k)}}\right\|}^2}}\ } 
\end{equation} 

$\left\| W^{\mathrm{(k)}}_{\mathrm{iteration}}- W^{\mathrm{(k)}}_{\mathrm{iteration-1}}\right\| \prec {\mathrm{I}}^{\mathrm{''}}_{\mathrm{k}}{\mathrm{I'}}_{\mathrm{k}}\mathrm{ \epsilon}$  the iterative process of MDA breaks, where $W^{\mathrm{(k)}}_{\mathrm{iteration}}$  $\mathrm{\in }$ ${\mathbb{ R}}^{{\mathrm{I}}^{\mathrm{''}}_{\mathrm{k} }{\mathrm{\times I'}}_{\mathrm{k}}}$. HOWSVD-MDA is detailed in  the algorithm ~\ref{Alg_1}.

\begin{algorithm*}
\caption{HOWSVD-MDA}\label{Alg_1}
%\State
\textbf{Inputs:}  
\begin{itemize}
 	
	\item M tensors $\mathrm{\boldsymbol{\mathrm{X}}\in {\mathbb{R}}^{I_1 \times I_2 \times \cdots \times I_N}}$ belonging to L diverse classes and each class j contains ${\mathrm{n}}_{\mathrm{j}}$ samples.
	
	\item  ${\mathrm{I'}}_1, {\mathrm{I'}}_2 , ..., {\mathrm{I'}}_{\mathrm{N}}$ the new lower dimensions after HOWSVD method.
	
	\item ${\mathrm{I''}}_1, {\mathrm{I''}}_2 , ..., {\mathrm{I''}}_{\mathrm{N}}$ the new lower dimensions after MDA method.
	\item $\mathrm{{itr}_{max}}$ is the maximal number of iterations.
\end{itemize}

%\State 
\textbf{Outputs:} 
\begin{itemize} 	
	\item $U^{\mathrm{(k)}}_{\mathrm{HOWSVD}} \in {\mathbb{R}}^{{\mathrm{I}}_{\mathrm{k}} \times {\mathrm{I'}}_{\mathrm{k}}}$ the projections matrices of HOWSVD method, k=1, 2,…, N. 
	\item $W^{\mathrm{(k)}}_{\mathrm{MDA}} \in {\mathbb{R}}^{{\mathrm{I'}}_{\mathrm{k}} \times {\mathrm{I''}}_{\mathrm{k}}}$ the projections matrices of MDA method, k=1, 2,…, N.

\end{itemize}

\begin{enumerate} 

	\item \textbf{For} k=1 to m

		\begin{enumerate} 

			\item $C^{\mathrm{(k)}}=\sum^{\mathrm{M}}_{\mathrm{i}} X^{\mathrm{(k)}}_{\mathrm{i}} (X^{\mathrm{(k)}}_{\mathrm{i}})^{\mathrm{T}}$.
			\item $ C^{\mathrm{(k)}}=U^{\mathrm{(k)}} \Sigma^{\mathrm{(k)}} {(V^{\mathrm{(k)}})}^\mathrm{T}$.
			\item $U^{\mathrm{(k)}}_{\mathrm{HOSVD}}= [u^{\mathrm{(k)}}_1, u^{\mathrm{(k)}}_2,..., u^{\mathrm{(k)}}_{\mathrm{N}}]$, where $U^{\mathrm{(k)}}_{\mathrm{HOSVD}}$ 	contains the selected left singular vectors corresponding to largest ${\mathrm{N}}_{\mathrm{k}}$ singular value. 
			\item $U^{\mathrm{(k)}}_{\mathrm{HOWSVD}}={(\Lambda^{\mathrm{(k)}})}^{-1/2} U^{\mathrm{(k)}}_{\mathrm{HOSVD}}, \Lambda^{\mathrm{(k)}}=[{\mathrm{\lambda}}^{\mathrm{(k)}}_1 {\mathrm{\lambda}}^{\mathrm{(k)}}_2 ...{\mathrm{\lambda}}^{\mathrm{(k)}}_{\mathrm{N}}]$. 

		\end{enumerate}

\item  ${\mathrm{\textbf{Y}}}_{\mathrm{i}}= {\mathrm{\textbf{X}}}_{\mathrm{i}} \times_1 {(U^{\mathrm{(1)}}_{\mathrm{HOWSVD}})}^{\mathrm{T}} \times_2 {(U^{\mathrm{(2)}}_{\mathrm{HOWSVD}})}^{\mathrm{T}}...\times_{\mathrm{N}} {(U^{\mathrm{(N)}}_{\mathrm{HOWSVD}})}^{\mathrm{T}}$.
		
\item \textbf{Initialization:} $W^{\mathrm{(1)}}_0=I_{{\mathrm{I'}}_{\mathrm{1}}}{,\ W}^{\mathrm{(2)}}_0=I_{{\mathrm{I'}}_{\mathrm{2}}},\dots ,W^{\mathrm{(N)}}_0=I_{{\mathrm{I'}}_{\mathrm{N}}}$.

\item \textbf{For } itr= 0 to $\mathrm{{itr}_{max}}$
		\begin{enumerate} 

			\item \textbf{For} k=1 to N
			
			\item ${\boldsymbol{\mathrm{Z}}}_{\mathrm{i}}={\boldsymbol{\mathrm{Y}}}_{\mathrm{i}}{\times }_{\mathrm{1}}W^{\mathrm{(1)}}_{\mathrm{itr}}\cdots {\times }_{\mathrm{N}}W^{\mathrm{(N)}}_{\mathrm{itr}} $.
			  \begin{itemize} 	
	
	\item $\boldsymbol{\mathrm{Z}}{\mathrm{\Rightarrow }}_{\mathrm{k}}Z^{\mathrm{(k)}}$.
 \item $S^{\mathrm{k}}_{\mathrm{b}}=\sum^{\mathrm{L}}_{\mathrm{j}=1}{{\mathrm{n}}_{\mathrm{j}}\left({\overline{Z}}^{\mathrm{k}}_{\mathrm{j}}-{\overline{Z}}^{\mathrm{k}}\right){\left({\overline{Z}}^{\mathrm{k}}_{\mathrm{j}}-{\overline{Z}}^{\mathrm{k}}\right)}^{\mathrm{T}}}$.
	\item $S^{\mathrm{k}}_{\mathrm{w}}=\sum^{\mathrm{L}}_{\mathrm{j}=1}{\sum^{{\mathrm{n}}_{\mathrm{j}}}_{\mathrm{i=1}}{{\left(Z^{\mathrm{k}}_{\mathrm{j,i}}-{\overline{Z}}^{\mathrm{k}}_{\mathrm{j}}\right)\left(Z^{\mathrm{k}}_{\mathrm{j,i}}-{\overline{Z}}^{\mathrm{k}}_{\mathrm{j}}\right)}^{\mathrm{T}}}}$;

\end{itemize}

\item \textbf{if} $itr \geq 2 $ and $\left\|W^{\mathrm{(k)}}_{\mathrm{iteration}}-W_{\mathrm{iteration-1}}^{\mathrm{(k)}}\right\| \prec {\mathrm{I}}^{\mathrm{''}}_{\mathrm{k}}{\mathrm{I'}}_{\mathrm{k}}\mathrm{\epsilon }$,

\textbf{break} after the convergence of the MDA : $W^{\mathrm{(k)}}_{\mathrm{MDA}} = W^{\mathrm{(k)}}_{\mathrm{itr}}$.

		%\end{enumerate}

\end{enumerate}

\end{enumerate}

\end{algorithm*}

\section{Knowledge pre-trained CNN based tensor subspace learning}
\label{sec:Knowpre}

\begin{figure*}[ht!]
\centering
\includegraphics[width=1\textwidth]{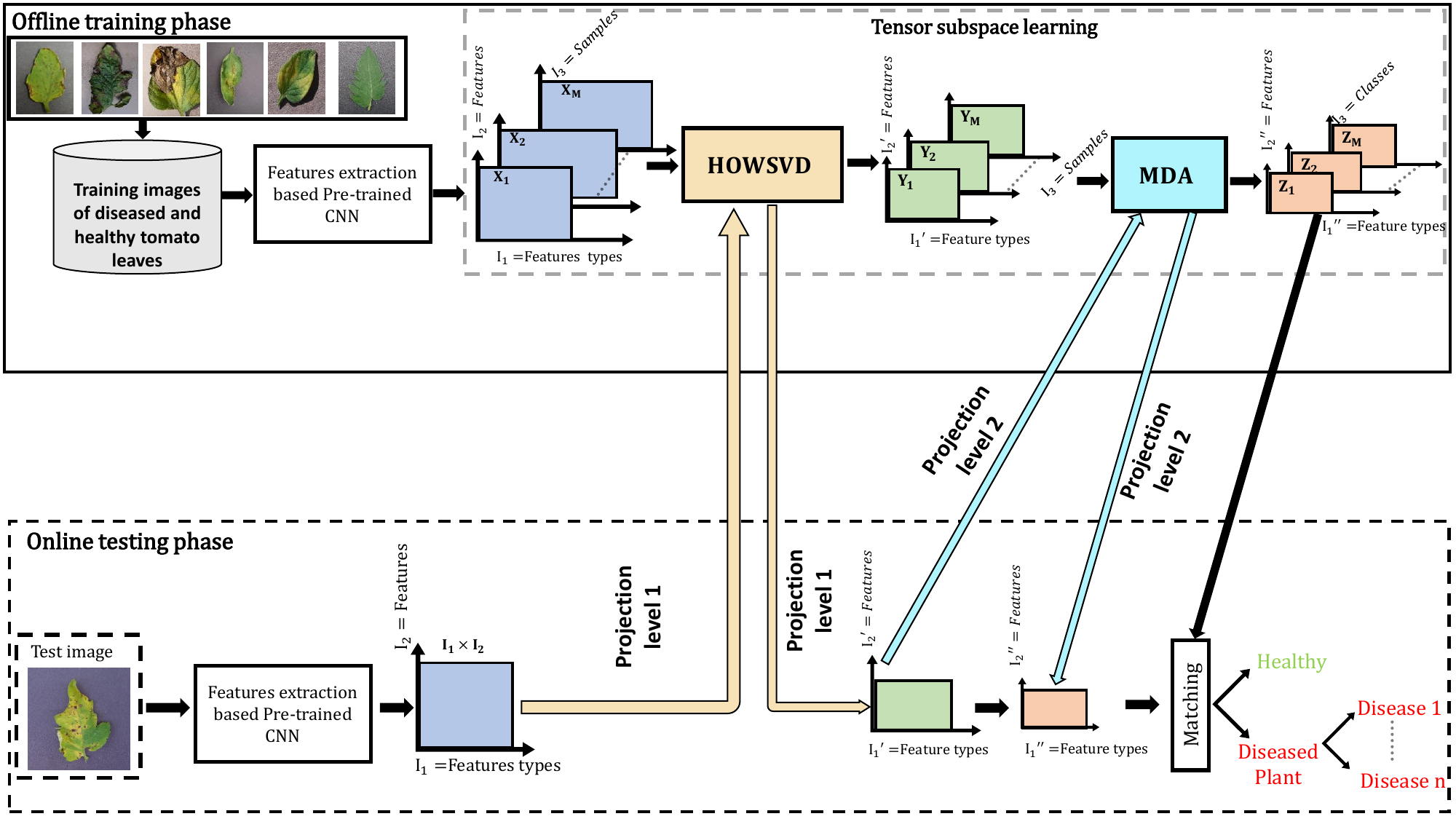}
\caption{Block diagram of the proposed system}
\label{fig3}
\end{figure*}

In this section, we provide detailed insights into the use of the proposed HOWSVD-MDA method for the classification of tomato leaf diseases. As illustrated in Fig.~\ref{fig3}, the block diagram of our approach comprises three integral components: knowledge-based pre-trained CNN, tensor subspace learning, and comparison. In this study, our primary emphasis is on the utilization of knowledge-based pre-trained CNN as a feature extraction strategy and the two-level tensor subspace learning, which is further elaborated in the following sections.

\subsection{Knowledge Pre-trained CNN}

The significance of large tomato image features in tomato leaf disease classification is well-established. Consequently, the use of deep features to elucidate leaf diseases becomes a critical issue and a significant challenge in the field of tomato leaf disease classification. To discover effective and discriminative features from tomato images, we propose to use prior knowledge and combine the deep features provided by multiple pre-trained CNN networks. We obtain a reduced set of deep features from the final fully connected (FC) layers in the CNN models, positioned just before the softmax layer. This subset contains a smaller number of deep features as compared to the earlier layers \cite{sharma2023novel,keceli2022deep,BELABBACI2023}.

In this work, we study and compare 18 species of pre-trained CNN models. Pre-trained CNNs come with a wealth of pre-acquired hierarchical features from extensive datasets. Using these pre-trained models as a foundation can notably decrease the quantity of labeled data needed for training, which is particularly advantageous when obtaining a substantial dataset poses difficulties. These pre-trained CNNs have proven their proficiency in capturing universal characteristics such as edges, textures, and elements of objects. By adapting these models with tomato leaf disease data, they can acquire specific disease-related patterns, leading to enhanced classification accuracy when contrasted with training a model from the ground up. Following this, we select the best-performing models and combine them using tensor subspace learning based on the proposed HOWSVD-MDA.
The proposed tensor subspace learning transfers the knowledge from large pre-trained CNNs to tomato leaf disease classification to find a metric space in which the tomato features become discriminative. Fig.~\ref{fig3} illustrates the proposed knowledge-based tensor subspace learning method. Each tomato image is characterized using multiple pre-trained deep feature descriptors extracted from the fully connected layer, which consists of 1000 neurons. These feature vectors for each tomato image are then represented as 2D arrays. Subsequently, all the images in the dataset are organized to create a third-order tensor data structure. Furthermore, tensor feature fusion has been incorporated into the process, where mode 1 represents various features, mode 2 signifies different feature types, and mode 3 is designated for the samples within the dataset. It is important to note that our simple tensor feature fusion strategy is employed without the use of conventional feature fusion techniques.

\begin{figure*}[ht!]
\centering
\includegraphics[width=1\textwidth]{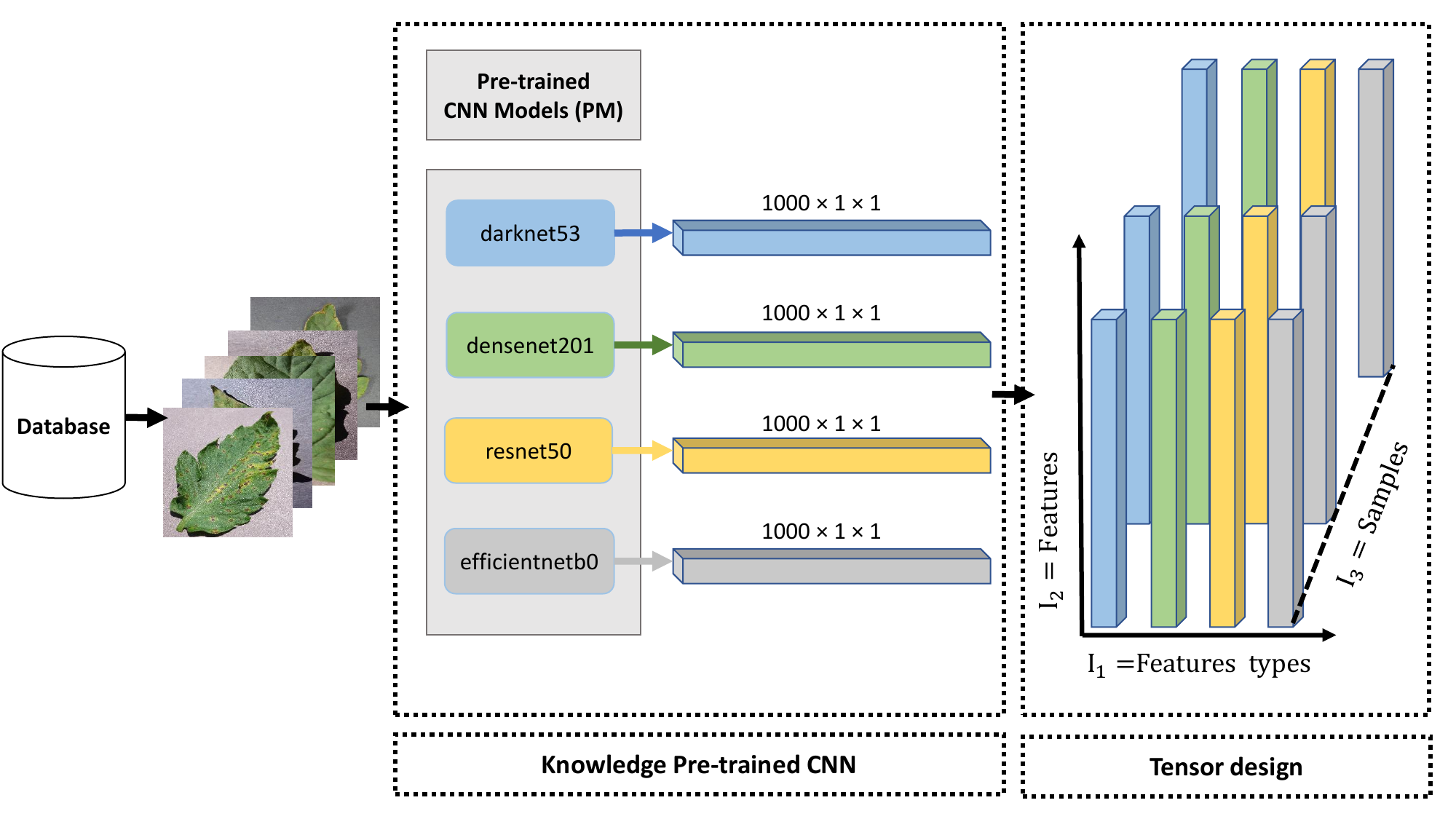}
\caption{Knowledge-based multi Pre-trained CNN features extraction and tensor design for tomato leaves images}
\label{fig4}
\end{figure*}

\subsection {Tensor subspace learning}
 
Each tomato image is represented using multiple pre-trained CNN models within mode 2 of the tensor data. The feature vectors of all training tomato images are organized into a third-order tensor $\mathrm{\boldsymbol{\mathrm{X}}\in {\mathbb{R}}^{I_1 \times I_2 \times I_3}}$, as illustrated in Fig.~\ref{fig4} in a block tensor design. In this context, $\mathrm{I_1}$ corresponds to a single feature vector from the pre-trained CNN ($\mathrm{I_1}$ = 1000), $\mathrm{I_2}$ represents different pre-trained CNN models, and $\mathrm{I_3}$ corresponds to the tomato image samples in the training database. This tensor ($\mathrm{\boldsymbol{\mathrm{X}}\in {\mathbb{R}}^{I_1 \times I_2 \times I_3}}$) is then subjected to projection using our proposed HOWSVD-MDA into a lower subspace $\mathrm{I'_1} \times \mathrm{I'_2}$, where $\mathrm{I'_1} \times \mathrm{I'_2} \prec \prec \mathrm{I_1} \times \mathrm{I_2}$. The reason for applying HOWSVD prior to MDA is to mitigate the issue of a small sample size problem (SSSP) across various tensor modes. This problem surfaces when the dimension of pre-trained CNN features exceeds the number of available samples, denoted as $M$, which can result in singular MDA scatter matrices. The initial step involves reducing the dimensionality of each tensor mode. This reduction in tensor dimensionality significantly enhances the effectiveness of the MDA method.

After applying HOWSVD, the training data tensor $\mathrm{\boldsymbol{\mathrm{Y}}\in {\mathbb{R}}^{I'_1 \times I'_2   \times I_3}}$ is projected using MDA to a new lower and discriminative subspace tensor $\mathrm{\boldsymbol{\mathrm{Z}}\in {\mathbb{R}}^{I''_1 \times I''_2   \times I_3}}$. For mode 3 ($\mathrm{I_3}$), we ignored the projection process since this dimension comprises the number of samples, whereas the learning of tomato leaf disease classification is based on these samples. Thus, we avoid the projection process in the third mode.

At the test phase, each tomato image is characterized as a 2$^{th}$-order tensor formed by stacking several pre-trained CNN models of the tomato image. Then, the tensor of dimension $\mathrm{I_1} \times \mathrm{I_2}$ is projected by HOWSVD-MDA, in which its dimensions become $\mathrm{I''_1} \times \mathrm{I''_2}$.

\subsection{Matching and Classification Decision}

In the matching phase of our study, we exploit the effectiveness of cosine similarity to achieve precise and efficient disease recognition. This process involves utilizing the reduced knowledge from pre-trained CNNs projected through the HOWSVD-MDA space, which are then concatenated to create a single feature vector. Subsequently, we employ cosine similarity \cite{Mohceneapp,6045519} to measure the similarity between the feature vector of the test tomato image and the model's feature vectors from the training dataset, resulting in a match score. This approach is further enhanced by the incorporation of discriminant analysis techniques, with cosine similarity distance offering distinct advantages due to its association with the Bayes decision rule \cite{OUAMANE201768}. Cosine similarity between two vectors $v_1$ and $v_2$ is defined as follows:

\begin{equation} \label{GrindEQ10} 
Cos(v_1, v_2 )= \frac{v_1^T . v_2}{\left\|v_1\right\| . \left\|v_2\right\|} 
\end{equation} 

Where $\left\|.\right\| $ is the Euclidean norm.

This comprehensive matching methodology ensures robust disease identification in our classification system.

After the calculation of similarity scores between the test tomato image and the training model projected by HOWSVD-MDA, the decision-making process is facilitated by selecting the maximum index among the obtained scores. This intuitive and efficient approach ensures that the proposed tomato leaf disease classification system makes confident decisions based on the most prominent match, thereby streamlining the recognition process and enhancing the overall accuracy of disease identification.

\section{Experiments and Results Analysis}
\label{sec:Experiments}

This section offers a practical overview of the implementation of our proposed approach along with a comprehensive analysis of the results. Section 5.1 presents a detailed dataset description, followed by an exploration of parameter settings in Section 5.2. The results of our tomato leaf disease classification approach are detailed in Section 5.3. In Section 5.4, we provide a comparative evaluation of our method's performance against existing methodologies.

\label{sec:results}

\subsection{Datasets Description}

\label{sec:Datasets}

In our quest to evaluate the proposed approach of tomato leaf disease classification, we have rigorously tested our approach on two diverse and challenging datasets: PlantVillage and Taiwan. In the following subsections, we provide detailed descriptions of these two datasets.

\subsubsection{PlantVillage}
The PlantVillage dataset \cite{hughes2016open} is a widely recognized collection of plant leaf images, paired with labels indicating the presence of diseases, if any. Employing this dataset to train our tomato plant disease classification model serves as a valuable means to showcase the model's efficiency in identifying different diseases. Additionally, it aids in setting a benchmark for comparing our approach with other methods. The dataset comprises 18,160 images of tomato plant leaves, categorized into ten groups based on their health status (healthy or diseased). These images are in the RGB color format and have dimensions of 256 × 256 pixels.

Table~\ref{tab:T2} provides the specific counts of training and test images for each class \cite{electronics12040827}.

\subsubsection{Taiwan Tomato Dataset}

The initial dataset for Taiwan tomato diseases \cite{thuseethan2022siamese} is relatively small, designed to assess the proposed model. It consists of 622 leaf samples, each with dimensions of 227 × 227 pixels, distributed across six distinct categories, including both healthy and various diseased classes. These classes include Bacterial spot (100 samples), Black mold (67 samples), Gray spot (84 samples), Healthy (106 samples), Late blight (98 samples), and powdery mildew (157 samples). 

In addition to this dataset, there is an extended dataset that incorporates more samples for each class. These additional samples are generated using a variety of augmentation techniques, such as clockwise rotation (at 90, 180, and 270-degree angles), mirroring (both horizontally and vertically), and adjustments to image brightness. Consequently, the final dataset comprises a total of 4,976 tomato leaf images, including augmented samples. In this dataset, we allocate 80\% for the training phase and 20\% for the testing phase, following the conventional practice in these systems.

\subsection{Parameter Settings}

Our assessment of performance follows the testing protocol outlined in \cite{electronics12040827} for the PlantVillage database, with Table~\ref{tab:T2} detailing the specific counts of training and testing images per class. For the Taiwan tomato dataset, we adopt the identical protocol as presented in \cite{thuseethan2022siamese}, allocating 80\% for the training phase and 20\% for testing. We investigate and compare 18 different species of pre-trained CNN models, including: 
Resnet101, mobilenetv2, inceptionv3, resnet50, nasnetlarge, Alexnet, vgg16, vgg19, darknet53, shufflenet, squeezenet, xception, densenet201, efficientnetb0, efficientnetb0, googlenet, inceptionresnetv2, resnet18. All of these pre-trained CNN are available at: \url{https://www.mathworks.com/help/deeplearning/ug/pretrained-convolutional-neural-networks.html}.
We resized the input tomato images to the required dimensions, for example, 224 × 224 for Resnet101, 256 × 256 for darknet53, and 224 × 224 for vgg16.

HOWSVD-MDA is set to 96\% energy of the eigenvalues, conserving the eigenvalues that deliver up to 96\% of the information. For example, in the Taiwan tomato dataset, the training tensor $\mathrm{\boldsymbol{\mathrm{X}}\in {\mathbb{R}}^{1000 \times 4 \times 3978}}$ provides the dimensions of the projections matrices:

$U^{\mathrm{(1)}}_{\mathrm{HOWSVD}}\in {\mathbb{R}}^{4 \times 3 }$, $U^{\mathrm{(2)}}_{\mathrm{HOWSVD}}\in {\mathbb{R}}^{1000 \times 900 }$, $W^{\mathrm{(1)}}_{\mathrm{MDA}}\in {\mathbb{R}}^{3 \times 2 }$, and $W^{\mathrm{(2)}}_{\mathrm{MDA}}\in {\mathbb{R}}^{900 \times 110 }$. To ensure the tracking of the convergence of the projection of the 3rd order tensor, a maximum iteration ($\mathrm{{itr}_{max}}$) limit of 5 is established based on empirical observations.

To showcase the experimental findings, we employ a widely recognized evaluation metric called accuracy. Accuracy effectively quantifies how closely the predicted tomato disease class aligns with the actual tomato disease class. The formula for calculating accuracy is provided below.

\begin{equation} \label{GrindEQ111} 
Accuracy= \frac{ {\mathrm{T}}_{\mathrm{p}} + {\mathrm{T}}_{\mathrm{n}} } { {\mathrm{T}}_{\mathrm{p}} + {\mathrm{T}}_{\mathrm{n}} + {\mathrm{F}}_{\mathrm{p}} + {\mathrm{F}}_{\mathrm{n}}  }
\end{equation}

Where: 
\begin{itemize}
    \item ${\mathrm{T}}_{\mathrm{p}}$ : True positive, representing correctly predicted positive images.
    \item ${\mathrm{T}}_{\mathrm{n}}$  : True negative, representing correctly predicted negative images.
    \item ${\mathrm{F}}_{\mathrm{p}}$ : False positive, representing incorrectly predicted positive images.
    \item ${\mathrm{F}}_{\mathrm{n}}$ : False negative, representing incorrectly predicted negative images.
\end{itemize}

\begin{table*}[t]
    \centering
        \caption{ Total number of training and test images are used for each class for PlantVillage datasets.}

    \begin{tabular}{c|c|c|c}
    \hline
        \textbf{Class} & \textbf{Training images} & \textbf{ Test images} & \textbf{Total images}  \\ \hline
        Healthy & 1122  & 469 & 1591  \\ 
        Bacterial spot  & 1585 & 542 & 2127  \\ 
        Early blight  & 766 & 234 & 1000  \\ 
        Late blight  & 1662 & 247 & 1909  \\ 
        Leaf mold  & 544 & 408 & 952  \\ 
        Septoria leaf spot  & 1306 & 465 & 1771  \\ 
        Spider mites two-spotted spider mite  & 1323 & 353 & 1676  \\ 
        Target spot  & 1335 & 69 & 1404  \\ 
        Tomato mosaic virus  & 269 & 104 & 373  \\ 
        Tomato yellow leaf curl virus  & 4974 & 383 & 5357  \\ \hline
    \end{tabular}
         \label{tab:T2}%

\end{table*}

\subsection{Discussion}

In this section, we showcase the experimental assessment of our method for classifying tomato leaf diseases and engage in a conversation regarding the achieved outcomes. We conduct various experiments to evaluate both the $2^{nd}$-order approach involving Linear Discriminant Analysis (LDA) \cite{6799229} and the $3^{rd}$-order tensor representations using our method HOWSVD-MDA. To ensure a fair comparison, we execute identical experiments following the same protocol on two datasets: PlantVillage and Taiwan. The experiments aim to analyze the performance of 18 different pre-trained CNN models. Utilizing pre-trained CNNs for classifying tomato leaf diseases offers several advantages, including enhanced performance, reduced data needs, faster training, access to cutting-edge architectures, and a reduced chance of overfitting, among other benefits. These advantages establish pre-trained models as a valuable resource for computer vision tasks such as disease classification. Regarding $2^{nd}$-order tensors, we explore the utilization of pre-trained CNNs without any further learning, alongside Linear Discriminant Analysis (LDA). In the context of 3$^{rd}$-order tensors, we delve into both HOSVD-MDA and our newly proposed HOWSVD-MDA to facilitate result comparisons.

\subsubsection{$2^{nd}$-order tensor approach}

Tables~\ref{tab:1} and~\ref{tab:3} present the classification accuracy of tomato leaf diseases using individual pre-trained CNN features within the two datasets, PlantVillage and Taiwan. In this analysis, we extracted a reduced set of deep features from the final fully connected (FC) layer of each pre-trained CNN. Indeed, two methods were employed: one without any metric learning (PM without ML) and another using Linear Discriminant Analysis (LDA) for feature transformation. The data is organized in the form of a $2^{nd}$-order tensor. The findings in Tables~\ref{tab:1} and~\ref{tab:3} reveal the following insights:

\textbf{Performance without Metric Learning (PM without ML):} The accuracy of tomato leaf diseases classification using pre-trained CNNs without any metric learning is relatively low with all models. Specifically, the classification accuracy does not exceed 74.93\% for the PlantVillage dataset and is even lower at 60.92\% for the Taiwan dataset. This suggests that relying solely on the features from the final fully connected layer of pre-trained CNNs is insufficient for accurate disease classification.

\textbf{Performance Improvement with LDA:} After applying LDA to the features extracted from pre-trained CNNs, the classification performance sees a significant improvement (more than 20\%). Discuss the role of LDA in dimensionality reduction and feature transformation, which helps in better separating the disease classes. These results underscore how LDA amplifies the discriminative capabilities of features extracted by pre-trained CNNs.

\textbf{Variation in results Among Different Pre-trained CNNs:} The obtained results exhibit variations depending on the specific pre-trained CNN model used. It is important to emphasize that certain CNN architectures outperform others in this context. These disparities in performance can be attributed to a range of factors, including distinctions in model architectures, learned features, and their suitability for transferability to the disease classification task.

\textbf{Identification of best pre-trained CNNs:} Densenet201 appears to be the best for the PlantVillage dataset with a high accuracy of 97.27\%, while efficientnetb0 is the top performer for the Taiwan dataset with an accuracy of 95.89%.

\begin{table}[!ht]
    \centering
    \caption{Performance of pretrained models on the Tomato PlantVillage dataset based on $2^{nd}$-order tensor approach (\%)}
    \begin{tabular}{c|c|c}
    \hline
        \textbf{CNN} & \textbf{PM without ML} & \textbf{LDA}  \\ \hline
        Resnet101 & 71.51 & \textbf{96.49}  \\ 
        mobilenetv2 & 73.60 & 89.80  \\ 
        inceptionv3 & 64.29 & 90.38  \\ 
        resnet50 & 70.38 & \textbf{96.24}  \\ 
        nasnetlarge & 60.21 & 90.52  \\ 
        Alexnet  & 69.76 & 93.10  \\ 
        vgg16 & 71.46 & 91.58  \\ 
        vgg19 & 63.32 & 89.49  \\ 
        darknet53 & 71.01 & \textbf{95.74}  \\ 
        shufflenet & 70.01 & 93.52  \\ 
        squeezenet & 41.12 & 65.15  \\ 
        xception & 68.10 & 90.80  \\ 
        densenet201 & 68.60 & \textbf{97.27}  \\ 
        efficientnetb0 & 69.10 & \textbf{95.80 } \\ 
        googlenet & 62.98 & 88.83  \\ 
        inceptionresnetv2 & 58.71 & 92.85  \\ 
        resnet18 & 74.93 & 91.44  \\ \hline
    \end{tabular}
     \label{tab:1}%
\end{table}

\begin{table}[!ht]
    \centering
     \caption{Performance of pretrained models on the Tomato taiwan with augmentation dataset  based on $2^{nd}$-order tensor approach (\%) }
    \begin{tabular}{c|c|c}
    \hline
        \textbf{CNN} & \textbf{PM without ML} & \textbf{LDA}  \\ \hline
        Resnet101 & 52.10 & \textbf{93.88}  \\ 
        mobilenetv2 & 55.11 & 87.57  \\ 
        inceptionv3 & 49.49 & 89.77  \\ 
        resnet50 & 49.79 & \textbf{95.09}  \\ 
        nasnetlarge & 51.40 & 91.68  \\ 
        Alexnet  & 60.92 & 63  \\ 
        vgg16 & 56.31 & 57.71  \\ 
        vgg19 & 56.21 & 55.11  \\ 
        darknet53 & 51.40 & \textbf{93.88}  \\ 
        shufflenet & 53.00 & 85.57  \\ 
        squeezenet & 52.70 & 84.96  \\ 
        xception & 49.79 & 90.98  \\ 
        densenet201 & 49.39 & \textbf{95.69}  \\ 
        efficientnetb0 & 54.80 & \textbf{95.89}  \\ 
        googlenet & 51.90 & 86.47  \\ 
        inceptionresnetv2 & 51.90 & 86.47  \\ 
        resnet18 & 50.90 & 81.86  \\ \hline
    \end{tabular}
    \label{tab:3}%
\end{table}

\subsubsection{3$^{rd}$-order tensor approach}

In this subsection, we delve into the results of our study, as detailed in Tables~\ref{tab:2} and ~\ref{tab:4}, focused on the classification of tomato leaf diseases. Our approach combines tensor subspace learning with pre-trained CNN models. Within these tables lie critical insights that we will analyze in depth, unveiling the broader implications of our findings:

\textbf{$3^{rd}$-vs. $2^{nd}$-order tensors:} The results obtained with both the PlantVillage and Taiwan datasets (see Tables ~\ref{tab:2} and ~\ref{tab:4}) unequivocally demonstrate that the $3^{rd}$-order tensor approach outperforms the $2^{nd}$-order tensor approach. This robustly validates our decision to employ high-order tensors for data modeling, underscoring their efficacy in the tomato diseases classification.

\textbf{Effect of whitening process:} A pivotal technique contributing to the enhancement of classification accuracy was the application of whitening. The comparison between HOSVD-MDA and HWOSVD-MDA clearly illustrates that our approach, which incorporates whitening, consistently yielded superior results across both datasets. The whitening process played a critical role in diminishing intra-class variability, a key factor in bolstering accuracy.

\textbf{Discrimination capabilities of tensors:} Our methodology made a significant improvement by applying tensor subspace learning. Notably, HWOSVD-MDA was able to extract discriminant information from the null space of the within-class scatter matrix for each k-mode of the tensor. This demonstrates that the method had strong discrimination capabilities compared to other approaches. Our tensor subspace learning strategy stands out for its capacity to preserve the inherent data structure by organizing feature vectors into a tensor representation. In contrast, linear transformations that arrange feature vectors in a linear fashion have been observed to lead to the loss of this natural data structure.

\textbf{Combination of pre-trained CNNs models:} Our study also combined knowledge from the best four pre-trained CNNs (darknet53, densenet201, resnet50, and efficientnetb0). This combination proved highly effective, achieving an impressive accuracy rate of 98.36\% for the PlantVillage database and 98.39\% for the Taiwan database. Our study further investigates the knowledge from the top-performing pre-trained CNNs, including Darknet53, Densenet201, ResNet50, and EfficientNetB0, through tensor feature fusion. This fusion proved to be remarkably effective, culminating in an outstanding accuracy rate of 98.36\% for the PlantVillage dataset and 98.39\% for the Taiwan dataset. In summary, this research highlights the efficacy of $3^{rd}$-order tensors, whitening, and tensor subspace learning in significantly enhancing the accuracy of tomato leaf disease classification. Furthermore, the combination of knowledge from multiple pre-trained CNNs has yielded exceptional results. These findings hold substantial value for the broader domain of tomato disease classification.

\begin{figure}[ht!]
\centering
\includegraphics[width=0.5\textwidth]{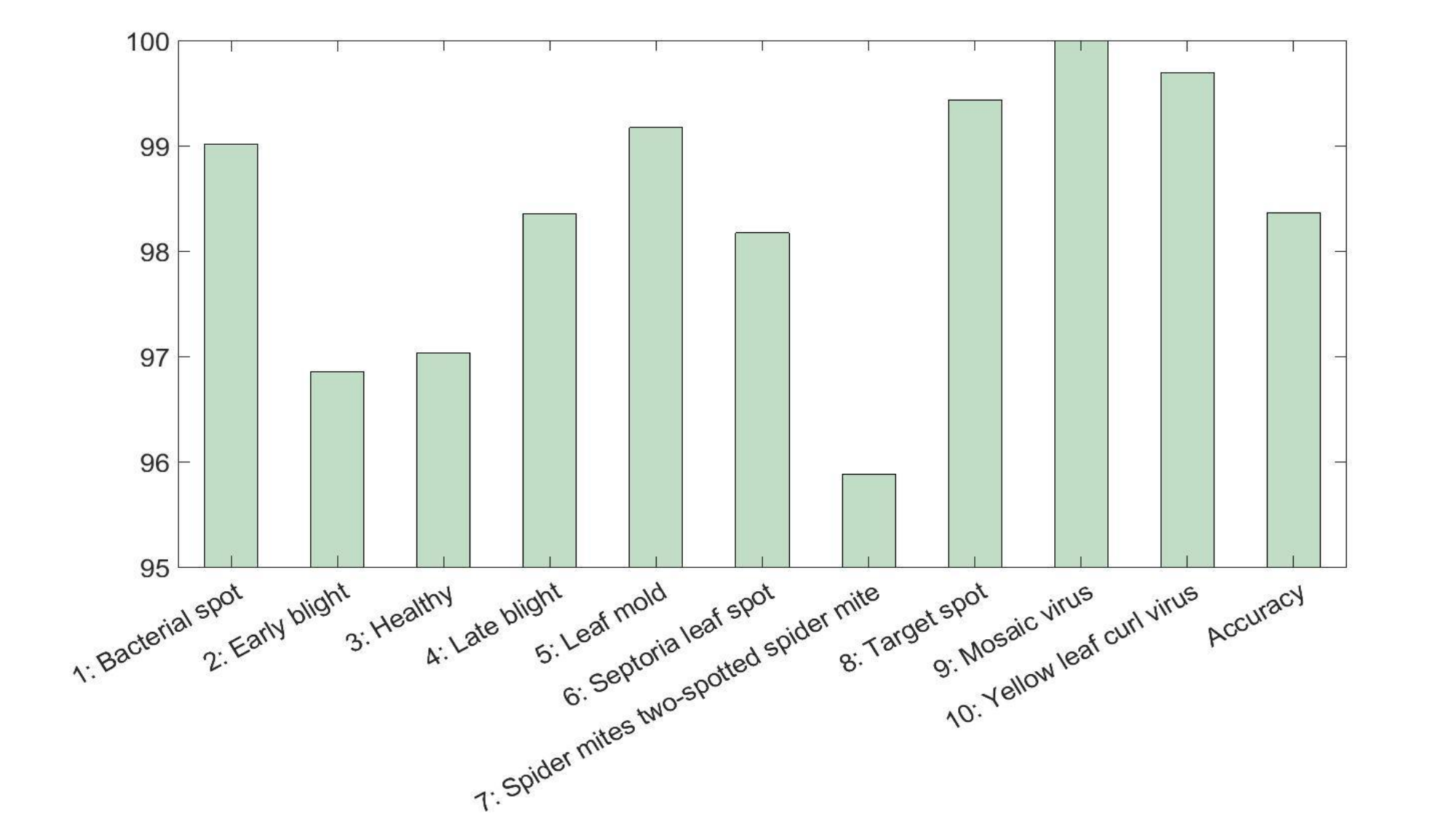}
\caption{Performance plot of the proposed HWOSVD-MDA for tomato plant leaf diseases on the Tomato PlantVillage dataset (\%)}
\label{fig11}
\end{figure}

\begin{figure}[ht!]
\centering
\includegraphics[width=0.5\textwidth]{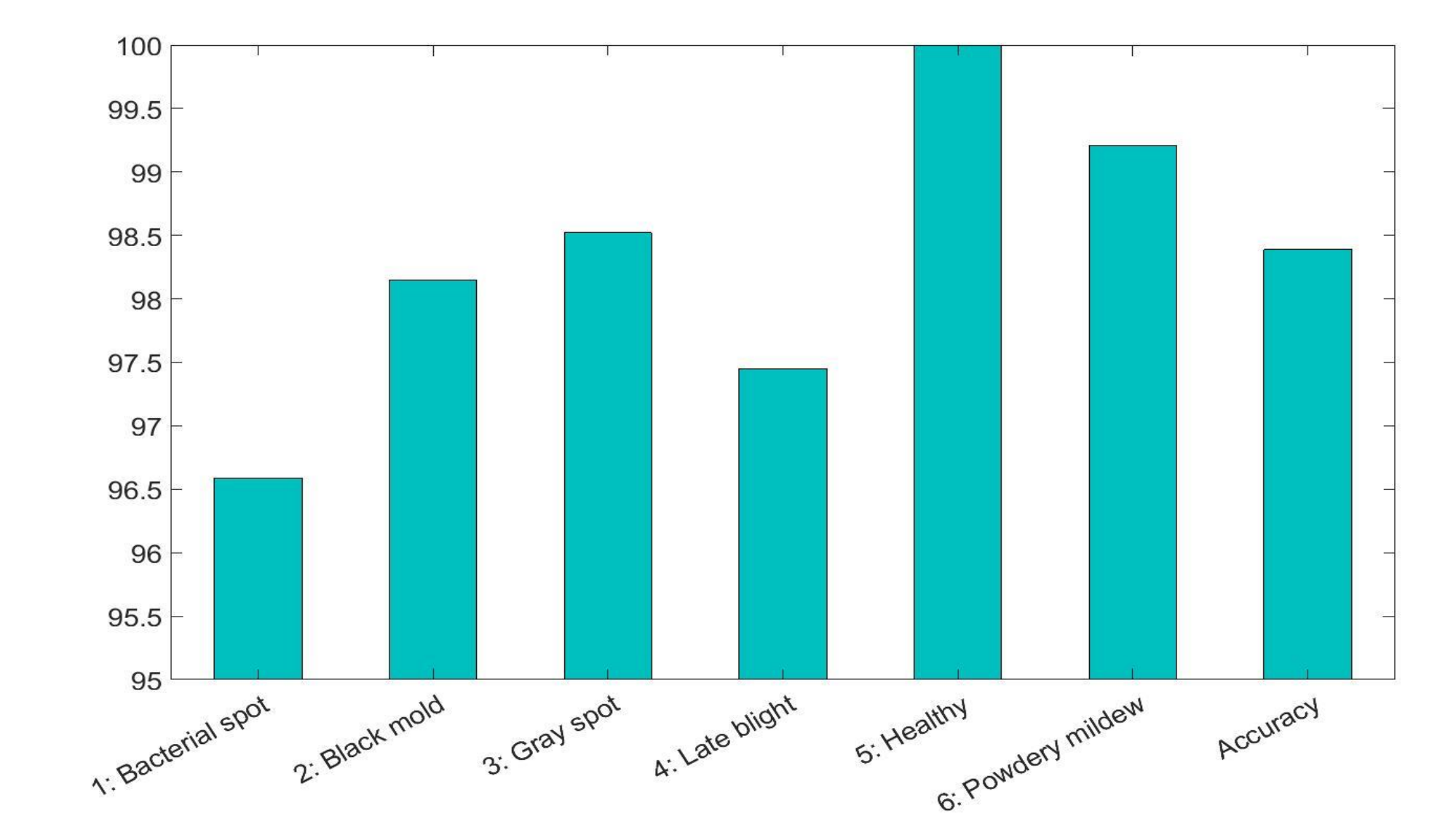}
\caption{Performance plot of the proposed HWOSVD-MDA for tomato plant leaf diseases on the Tomato Taiwan dataset with augmentation (\%)}
\label{fig12}
\end{figure}

\begin{figure}[ht!]
\centering
\includegraphics[width=0.5\textwidth]{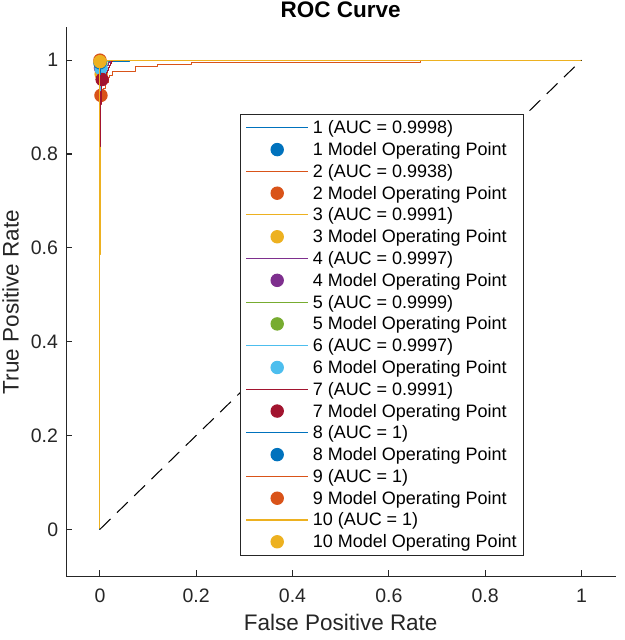}
\caption{ ROC graph of the proposed HWOSVD-MDA on the Tomato PlantVillage dataset (\%)}
\label{ROC1}
\end{figure}

\begin{figure}[t!]
\centering
\includegraphics[width=0.5\textwidth]{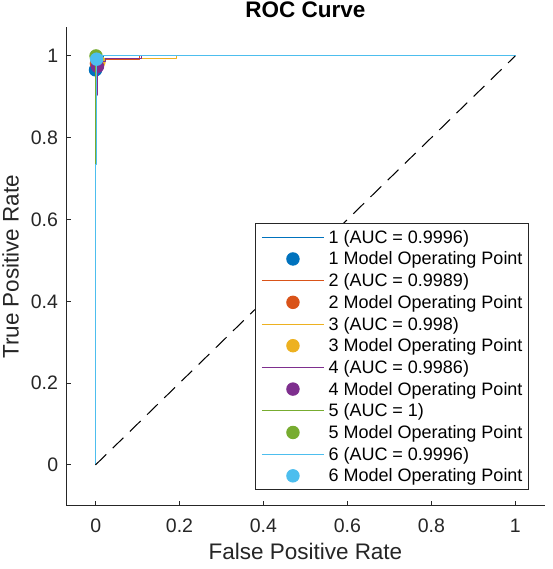}
\caption{ ROC graph of the proposed HWOSVD-MDA on Tomato Taiwan with augmentation dataset (\%)}
\label{ROC2}
\end{figure}

\textbf{Performance of the proposed HWOSVD-MDA for various categories of tomato plant leaf diseases:}
The outcomes for various categories of tomato plant leaf diseases achieved by the proposed HWOSVD-MDA approach are illustrated separately for Tomato PlantVillage and Tomato Taiwan with augmentation datasets in Figs.~\ref{fig11} and \ref{fig12}. Analyzing these figures reveals that the performance results are presented on a per-class basis. Furthermore, it is observed that each class exhibits values closely aligned with the overall success rate. The ROC and AUC values in Figs. \ref{ROC1} and \ref{ROC2} generally align closely to one another. Upon evaluating these metrics, it becomes apparent that variations in sample sizes for individual classes do not have an adverse impact on the study's results.

\begin{table*}[!ht]
    \centering
    \caption{Performance of pretrained models fusion using the proposed  HWOSVD-MDA on the Tomato PlantVillage dataset (\%) }
    \begin{tabular}{c|c|c|c}
    \hline
       \textbf{ Tensor order} & \textbf{Method} & darknet53+densenet201+resnet101 & darknet53+densenet201 +resnet50+ efficientnetb0  \\ \hline
        \textbf{$2^{nd}$-order tensor} & PM without ML & 75.79 & 76.74  \\ 
        ~ & LDA & 96.75 &  97.33 \\ \hline
       \textbf{ 3$^{rd}$-order tensor} & HOSVD-MDA & 97.01 & 97.91  \\ 
        ~ & \textbf{HWOSVD-MDA} & 97.66 & \textbf{98.36}  \\ \hline
    \end{tabular}
     \label{tab:2}%
\end{table*}

\begin{table*}[!t]
    \centering
     \caption{Performance of pretrained models fusion using the proposed  HWOSVD-MDA on the Tomato Taiwan with augmentation dataset  (\%) }
    \begin{tabular}{c|c|c|c}
    \hline
       \textbf{ Tensor order} & \textbf{Method} & darknet53+densenet201+resnet101 & darknet53+densenet201 +resnet50+ efficientnetb0  \\ \hline
        \textbf{$2^{nd}$-order tensor}  & PM without ML & 52.80 & 53.40  \\ 
        ~ & LDA & 95.20 & 95.67  \\ \hline
        \textbf{ 3$^{rd}$-order tensor} & HOSVD-MDA & 96.34 & 96.68  \\ 
        ~ & \textbf{HWOSVD-MDA} & 97.39 & \textbf{98.39}  \\ \hline
    \end{tabular}
     \label{tab:4}%
\end{table*}

\subsection{Comparison with state-of-art methods}

Furthermore, we evaluated the performances of our approach by comparing it to the state-of-the-art methods listed in Table ~\ref{tab:5} for the PlantVillage dataset and Table ~\ref{tab:6} for the Taiwan dataset with augmentation. It is worth noting that the results in Table ~\ref{tab:6} and those found in \cite{AGARWAL2020293,chen2020identification,trivedi2021early,agriculture12020228,astani2022diverse} for the compared methods were not directly extracted from their original papers due to differences in experimental conditions. Therefore, we re-executed these methods on the tomato datasets within a consistent environment to facilitate a meaningful comparison.

Furthermore, we conducted a thorough performance evaluation by comparing our approach HWOSVD-MDA to the state-of-the-art methods, as documented in Table ~\ref{tab:5} for the PlantVillage dataset and Table ~\ref{tab:6} for the Taiwan dataset with augmentation. It is noteworthy that the results in Table ~\ref{tab:6} and those reported in the compared methods \cite{AGARWAL2020293,chen2020identification,trivedi2021early,agriculture12020228,astani2022diverse} were not directly extracted from their original papers due to several variations in experimental conditions. To ensure a fair comparison, we re-implemented these methods on the tomato datasets within a consistent experimental environment.

Based on prior previous experiments, the optimal performance observed for tomato leaf diseases classification was achieved by using four knowledge pre-trained CNN models (Darknet53, Densenet201, Resnet50, and Efficientnetb0) and applying our tensor subspace learning (HWOSVD-MDA) method. The accuracy of 98.36\% and 98.39\% on the PlantVillage and Taiwan datasets, respectively. From these tables, we can see that our method surpasses the current state-of-the-art in all scenarios, highlighting the efficacy of using tensor-based tomato image representation. Tensors inherently manage multidimensional data, simplifying its manipulation, transformation, and feature extraction.

Our most promising results for tomato leaf disease classification emerged when we leveraged the knowledge from four pre-trained CNN models (Darknet53, Densenet201, ResNet50, and EfficientNetB0), coupled with our tensor subspace learning approach (HWOSVD-MDA). This combination delivered remarkable accuracy rates of 98.36\% for the PlantVillage dataset and 98.39\% for the Taiwan dataset. The tables ~\ref{tab:5} and ~\ref{tab:6} clearly demonstrate that our method outperforms the current state-of-the-art across all scenarios, underscoring the effectiveness of employing tensor-based representations for tomato images. Tensors inherently simplify the handling, transformation, and feature extraction of multidimensional data, further solidifying their utility in this domain.

\begin{table}[!ht]
    \centering
    \caption{Comparison of existing tomato leaf disease recognition methods using PlantVillage database   }
    \begin{tabular}{c|c|c}
    \hline
        \textbf{Authors} & \textbf{Year}   & \textbf{Accuracy (\%)}  \\ \hline 
        Agarwal et al.  \cite{AGARWAL2020293} & 2020 &  88.63  \\  
        Chen et al. \cite{chen2020identification}  & 2020 &  90.12  \\  
        Trivedi et al.  \cite{trivedi2021early}   & 2021 &  84.63  \\  
        Amreen et al.  \cite{abbas2021tomato}  & 2021 &  97.11  \\  
        Bhujel et al. \cite{agriculture12020228}  & 2022 &  91.36  \\  
        Astani et al.  \cite{astani2022diverse}  & 2022 &  93.21  \\  
        Thuseethan et al.  \cite{thuseethan2022siamese}  & 2022 &  96.97  \\  
        Moussafir et al.  \cite{moussafir2022design}  & 2022 &  98.10  \\  
        Zhao et al.  \cite{zhao2022ric}  & 2022 &  98.00  \\ 
        Paymode et al.  \cite{PAYMODE202223}  & 2022 &  95.71  \\  
        Laura et al. \cite{FALASCHETTI2022e00363}  & 2022 & 96:24  \\ 
        Lingwal et al.  \cite{10.1007/978-981-19-5868-7_15}  & 2023 &  83.00 \\   
        \textbf{Ours} &  \textbf{2023}  & \textbf{98.36}  \\ \hline
    \end{tabular}
 \label{tab:5}%
   \end{table} 

    \begin{table}[!ht]
    \centering
    \caption{Comparison of existing tomato leaf disease recognition methods using Taiwan with augmentation database   }
    \begin{tabular}{c|c|c}
    \hline
           \textbf{Authors} & \textbf{Year}  & \textbf{Accuracy (\%)}  \\ \hline 
        Agarwal et al.  \cite{AGARWAL2020293} & 2020 &  83.68  \\  
        Chen et al.  \cite{chen2020identification} & 2020 &  86.72  \\  
        Trivedi et al.  \cite{trivedi2021early}  & 2021 &  74.63  \\  
        Bhujel et al.  \cite{agriculture12020228} & 2022 &  88.92  \\  
        Astani et al.  \cite{astani2022diverse} & 2022 &  91.32  \\  
        Thuseethan et al. \cite{thuseethan2022siamese} & 2022 &  96.14  \\  
 
        \textbf{Ours} & \textbf{2023} &  \textbf{98.39}  \\ \hline
    \end{tabular}
     \label{tab:6}%
\end{table}

% conclusion.tex
\section{Conclusion}
\label{sec:conclusion}
In this research, we have introduced an innovative approach for Tomato Leaf Disease classification. Our proposed method, named HOWSVD, performs multilinear subspace learning of multidimensional data. This approach combines Tensor Subspace Learning with the integration of knowledge-driven pre-trained CNN features to enhance the discriminative power of the system. These essential features are then projected into a novel tensor-based discriminative subspace using a two-level approach based on HOWSVD-MDA. The evaluation encompassed two distinct datasets, and the results unequivocally demonstrate the superior performance of HOWSVD-MDA when compared to alternative methods, underscoring the potential of a tensor-based approach in this context. The success achieved with the Knowledge Pre-trained CNNs tensor representation encourages further exploration and underscores the significance of HOWSVD-MDA for future research, particularly in domains such as Federated Learning-based IoT applications.

Moreover, the research provides a valuable contribution to the field by showcasing a method that can adapt to various datasets (PlantVillage and Taiwan) and maintain high performance across different conditions. The success of this approach in achieving high classification accuracy rates (98.36\% for PlantVillage and 98.39\% for Taiwan) sets a new benchmark for the state-of-the-art in tomato leaf disease classification and highlights the potential for further exploration and application of tensor-based machine learning models in agriculture and beyond.

This future work not only aims to build upon the findings of the current study but also to address broader challenges in agricultural technology, plant disease management, and the application of machine learning in real-world scenarios.


\begin{IEEEbiography}[{\includegraphics[width=1in,height=1.25in,clip,keepaspectratio]{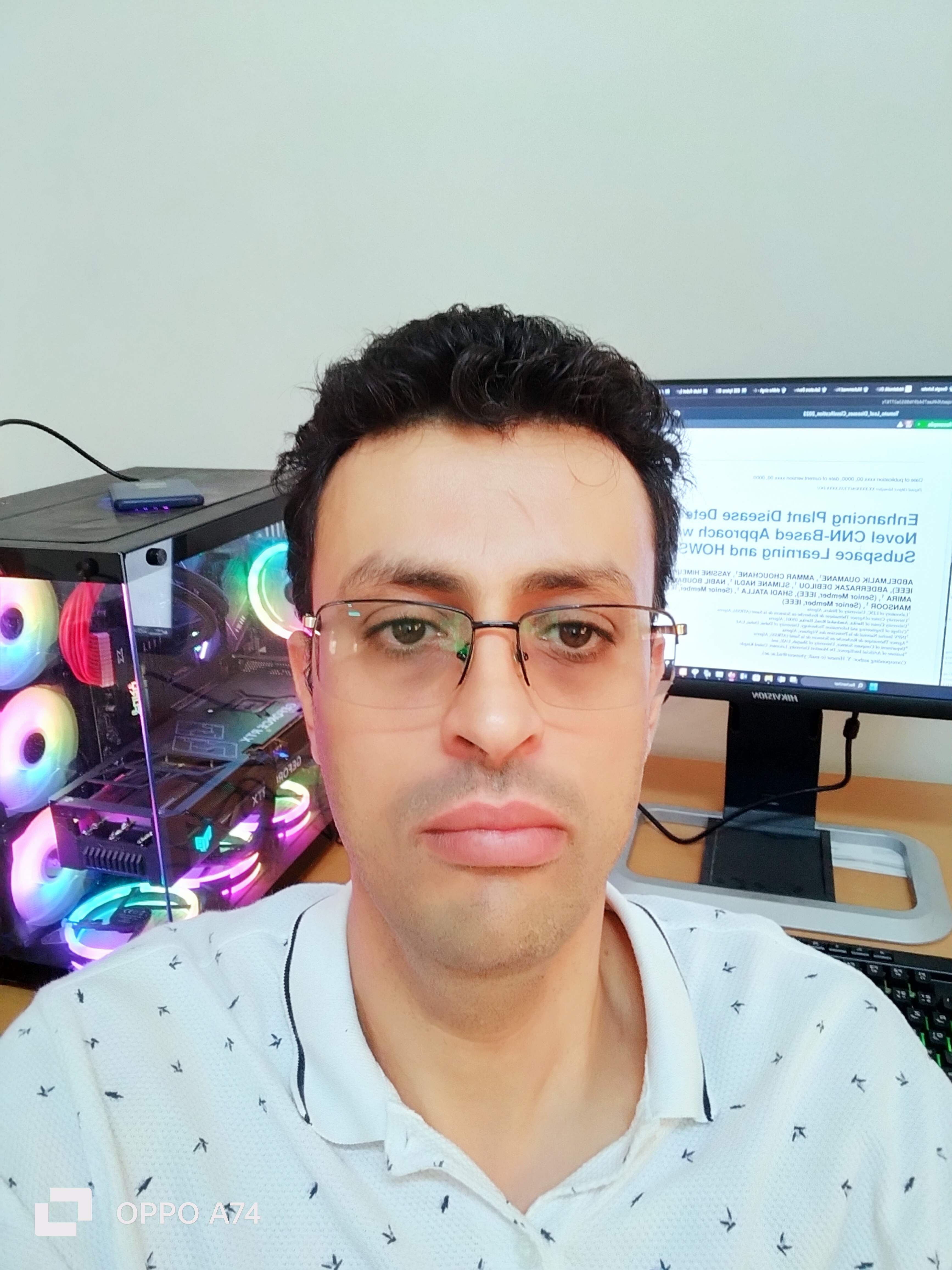}}]{Prof. Abdelmalik Ouamane} received the Ph.D. degree in Electronics from the University of Biskra in 2015. He is currently a Professor with Department of Electrical Engineering, University of Biskra, Algeria. He has been a reviewer for many conferences and journals. His research interests include Computer vision, pattern recognition, machine learning, deep learning, biometrics, soft biometrics, development of efficient methods for tensor analysis and plant sciences.
\end{IEEEbiography}

\begin{IEEEbiography}[{\includegraphics[width=1in,height=1.25in,clip,keepaspectratio]{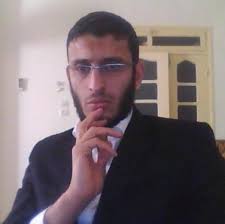}}]{Dr. Ammar Chouchane} is a distinguished researcher who obtained his PhD degree in the field of telecommunications from the University of Biskra, Algeria, in June 2016. Currently, he serves as an Associate Professor at the University Centre of Barika, Algeria. Dr. Ammar has made significant contributions to various domains, including pattern recognition, computer vision, biometrics, and image processing. Additionally, his expertise extends to emerging technologies such as federated learning and smart agriculture. 
\end{IEEEbiography}

\begin{IEEEbiography}[{\includegraphics[width=1in,height=1.25in,clip,keepaspectratio]{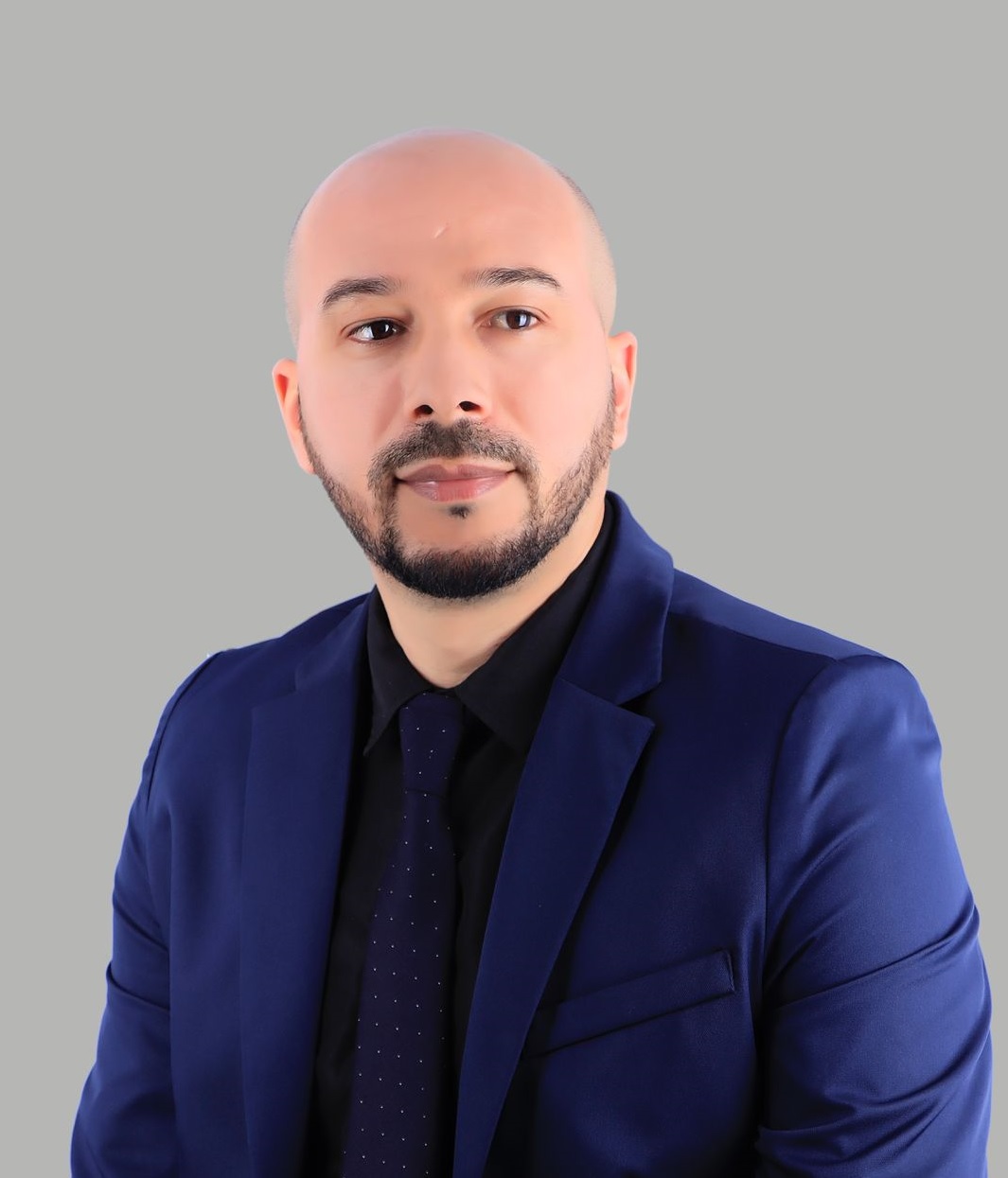}}]{Dr. Yassine Himeur (Senior Member, IEEE)} is presently an Assistant Professor of Engineering \& Information Technology at the University of Dubai. He completed both his M.Sc. and Ph.D. degrees in Electrical Engineering in 2011 and 2015, respectively. Following his doctoral studies, he obtained the Habilitation to Direct Research, which granted him the official authorization to supervise research, in July 2017. His academic journey led him to join the faculty at the University of Dubai after serving as a Postdoctoral Research Fellow at Qatar University from 2019 to 2022. Prior to that, he held the position of Senior Researcher from 2013 to 2019 at the Algerian Center for Development of Advanced Technologies (CDTA), where he also served as the Head of the TELECOM Division from 2018 to 2019. Throughout his career, he has been actively involved in conducting R\&D projects and has played a significant role in proposing and co-leading several research proposals under the NPRP grant (QNRF, Qatar). With more than 150 research publications in high-impact venues, he has made valuable contributions to the field. He was honored to receive the Best Paper Award at the 11th IEEE SIGMAP in Austria in 2014 and Best Student Paper Award at IEEE GPECOM 2020 in Turkey. His current research interests encompass Big Data and IoTs, AI/ML/DL, Healthcare Technologies, Recommender Systems, Building Energy Management, and Cybersecurity.
\end{IEEEbiography}

\begin{IEEEbiography}[{\includegraphics[width=1in,height=1.25in,clip,keepaspectratio]{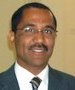}}]{Prof. Abderrazak Debilou} received the Ph.D. degree in Instrumentation et mesures from the University of 	Bordeaux 1 in 1990. He is currently a Professor with Department of Electrical Engineering, University of Biskra, Algeria. His research interests include Computer vision, machine learning, instrumentation and measurements. 
\end{IEEEbiography}

\begin{IEEEbiography}[{\includegraphics[width=1in,height=1.25in,clip,keepaspectratio]{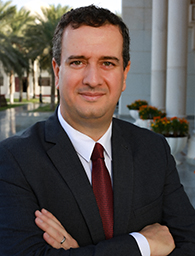}}]{Prof. Abbes Amira (Senior Member, IEEE)} received the Ph.D. degree in the area of computer engineering from Queen’s University, Belfast, U.K., in 2001. He is currently the Embedded Computing Research Co-Ordinator with Qatar University. He took many academic and consultancy positions, including his current position as a Professor in computer engineering and the acting Director of the KINDI Center for computing research with Qatar University. His research interests include reconfigurable computing, signal processing and connected health. He is a fellow of the IET, the HEA, and the ACM.
\end{IEEEbiography}

\begin{IEEEbiography}[{\includegraphics[width=1in,height=1.25in,clip,keepaspectratio]{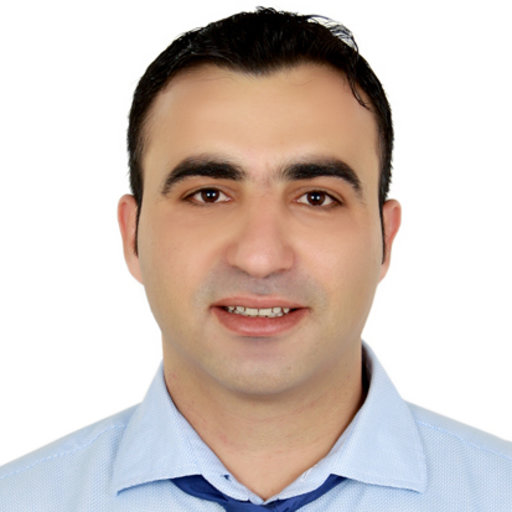}}]{Dr. Shadi Atalla (Member, IEEE)} is an Associate Professor and Director of the Computing \& Information Systems program at the University of Dubai. With over 15 years of experience in teaching and research, he is a prominent data science evangelist and certified big data trainer, highly regarded in the industry. Dr. Atalla research focuses on developing data science algorithms, curriculum development, and artificial intelligence. He has published several papers in international scientific journals and has contributed significantly to the field of data science. Dr. Atalla also serves as the Chair of the Computer Society of IEEE UAE, showcasing his leadership qualities and dedication to advancing the field.  
\end{IEEEbiography}

\begin{IEEEbiography}[{\includegraphics[width=1in,height=1.25in,clip,keepaspectratio]{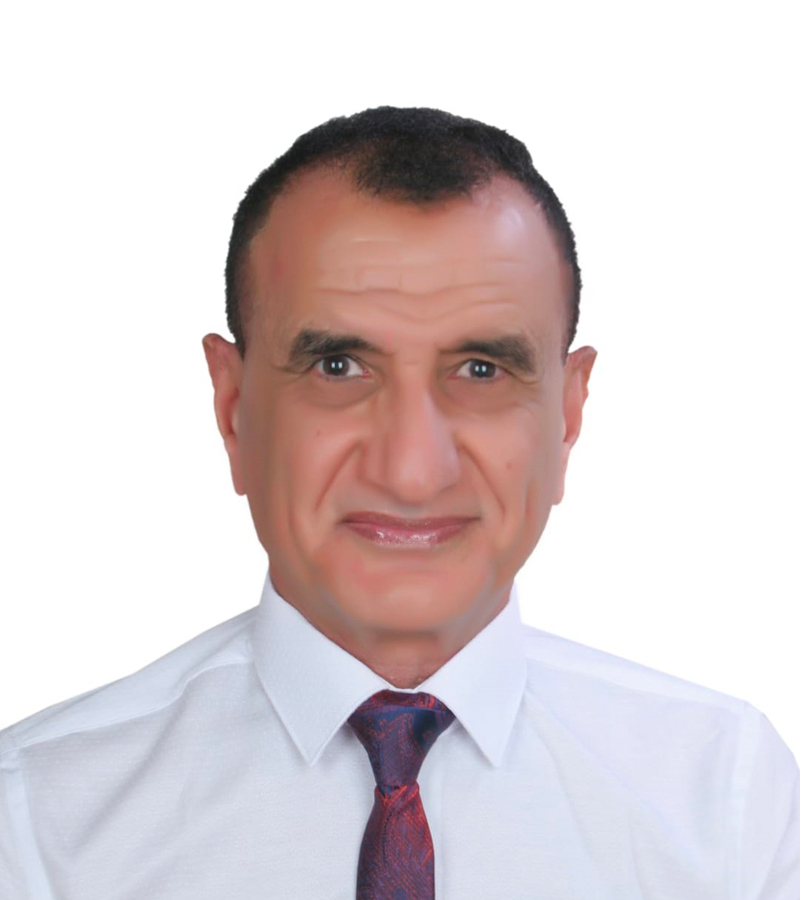}}]{Prof. Wathiq Mansoor (Senior Member, IEEE)} is a Professor at University of Dubai. He has an excellent academic leadership experience in well-known universities worldwide. He earned his Ph.D. in computer engineering from Aston University in UK. His doctoral work was on the design and implementations of multiprocessors systems and communications protocols for computer vision applications. He has published many research papers in the area of Intelligent Systems, Image processing, deep learning, Security, ubiquitous computing, web services, and neural networks. His current research is in the area of intelligent systems and security using neural networks with deep learning models for various applications. He has organized many international and national conferences and workshops. He is a senior member of IEEE UAE section. He has supervised many Ph.D. and undergraduate projects in the field of Computer engineering and innovation in business, in addition to co-supervise many postgraduate students  through research collaboration with international research groups.  
\end{IEEEbiography}

\begin{IEEEbiography}[{\includegraphics[width=1in,height=1.25in,clip,keepaspectratio]{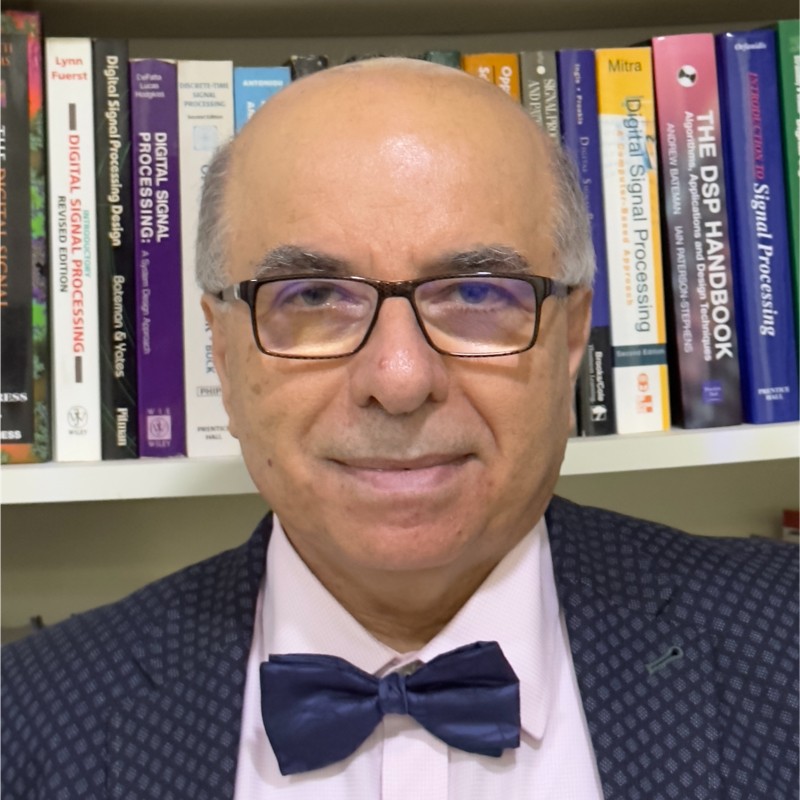}}]{Prof. Hussain Al-Ahmad (Life Senior Member, IEEE)} received his Ph.D. from the University of Leeds, UK in 1984 and currently he is the Provost and Chief Academic Officer at the University of Dubai, UAE. He has 37 years of higher education experience working at academic institutions in different countries including University of Portsmouth, UK, Leeds Beckett University, UK, Faculty of Technological Studies, Kuwait, University of Bradford, UK, Etisalat University College, Khalifa University and University of Dubai, UAE. He was the founding Dean of Engineering and IT at the University of Dubai, UAE. He was the founder and Chair of the Electronic Engineering department at both Khalifa University and Etisalat University College. His research interests are in the areas of signal and image processing, artificial intelligence, remote sensing and propagation. He has supervised successfully 32 PhD and Master students in the UK and UAE. He has delivered short courses and seminars in Europe, Middle East and Korea. He has published over 100 papers in international conferences and journals. He has UK and US patents. He served as chairman and member of the technical program committees of many international conferences. He is a Life Senior Member of the IEEE and a Fellow of many prestigious institutes. He is currently the Chair of the IEEE UAE Section.  
\end{IEEEbiography}

\EOD

\end{document}